\documentclass[a4paper,fleqn]{cas-dc}

\usepackage[authoryear,longnamesfirst]{natbib}

\begin{document}
\let\WriteBookmarks\relax
\def\floatpagepagefraction{1}
\def\textpagefraction{.001}

\newtheorem{example}{Example}
\newtheorem{theorem}{Theorem}

\shorttitle{Weakly Supervised Label Learning Flows}

%\shortauthors{CV Radhakrishnan et~al.}

\title [mode = title]{Weakly Supervised Label Learning Flows}                      
% \tnotemark[1,2]

% \tnotetext[1]{This document is the results of the research
%    project funded by the National Science Foundation.}

% \tnotetext[2]{The second title footnote which is a longer text matter
%    to fill through the whole text width and overflow into
%    another line in the footnotes area of the first page.}

\author[1]{You Lu}
% [orcid=0000-0002-4357-1888]

\fnmark[1]
\fntext[1]{Work was done while You Lu was studying in Virginia Tech.}

% \cormark[1]


\ead{youlu1206@gmail.com}

% \ead[url]{www.anonymous.authors.com}

\credit{Conceptualization, Methodology, Software, Writing, Review, Edit}

\affiliation[1]{organization={Motional},
    addressline={100 Northern Ave Suite 200}, 
    city={Boston},
    postcode={02210}, 
    country={U.S.}}

\author[2]{Wenzhuo Song}
%[orcid=0000-0002-1607-3402]

\cormark[2]
\cortext[2]{Corresponding author}


\ead{wzsong@nenu.edu.cn}

% \ead[url]{www.anonymous.authors.com}

\credit{Writing, Review, Edit}

\affiliation[2]{organization={Northeast Normal University},
    addressline={2555 Jingyue Street}, 
    city={Changchun},
    % citysep={}, % Uncomment if no comma needed between city and postcode
    postcode={130117}, 
    country={China}}

\author[3]{Chidubem Arachie}
% []

% \cormark[3]

% \fnmark[3]

\ead{achid17@vt.edu}

% \ead[url]{www.anonymous.authors.com}

\credit{Review, Edit}

\affiliation[3]{organization={Google},
    addressline={1195 Borregas Drive}, 
    city={Sunnyvale},
    postcode={94089}, 
    country={U.S.}}

\author[4]{Bert Huang}
%[orcid=0000-0002-8548-7246]

\ead{berthuang@gmail.com }

\credit{Administration, Supervision, Review, Edit}

\affiliation[4]{organization={Snorkel AI},
    addressline={1178 Broadway}, 
    city={New York},
    postcode={10001}, 
    country={U.S.}}

\begin{abstract}
Supervised learning usually requires a large amount of labelled data. However, attaining ground-truth labels is costly for many tasks. Alternatively, weakly supervised methods learn with cheap weak signals that only approximately label some data. Many existing weakly supervised learning methods learn a deterministic function that estimates labels given the input data and weak signals. In this paper, we develop label learning flows (LLF), a general framework for weakly supervised learning problems. Our method is a generative model based on normalizing flows. The main idea of LLF is to optimize the conditional likelihoods of all possible labelings of the data within a constrained space defined by weak signals. We develop a training method for LLF that trains the conditional flow inversely and avoids estimating the labels. Once a model is trained, we can make predictions with a sampling algorithm. We apply LLF to three weakly supervised learning problems. Experiment results show that our method outperforms many baselines we compare against.
\end{abstract}

% \includegraphics{figs/grabs.pdf}

% \begin{highlights}
% \item Research highlights item 1
% \item Research highlights item 2
% \item Research highlights item 3
% \end{highlights}

% Keywords

\maketitle

\section{Introduction}
\label{sec:introduction}
Machine learning has achieved great success in many supervised learning tasks. However, in practice, data labeling is usually human intensive and costly. To address this problem, practitioners are turning to weakly supervised learning~\citep{zhou2018brief}, which trains machine learning models with only noisy labels generated by human specified rules or pretrained models. Currently, weakly supervised learning has been applied to many practical problems~\citep{ratner2017snorkel,bach2019snorkel,fries2019weakly}.

In this paper, we focus on a new line of research--constraint-based weakly supervised learning. Existing methods~\citep{balsubramani2015scalable,arachie2020constrained,arachie2021general,mazzetto:icml21,mazzetto2021semi} learn deterministic functions that estimate unknown labels $\mathbf{y}$ for given input data $\mathbf{x}$ and weak signals $\mathbf{Q}$. Since the observed information is incomplete, the predictions based on it can be varied. That is, an input $\mathbf{x}$ can be corresponding to multiple possible output $\mathbf{y}$. However, current methods ignore this uncertainty between input and output. To address this problem, we develop \emph{label learning flows} (LLF) \footnote{code is available at: \href{https://github.com/yolu1055/weakly-supervised-label-learning-flow}{https://github.com/yolu1055/weakly-supervised-label-learning-flow}}, which is a general framework for weakly supervised learning problems. Our method models the uncertainty between $\mathbf{x}$ and $\mathbf{y}$ with a probability distribution $p(\mathbf{y}|\mathbf{x})$. We use a conditional generative flow~\citep{dinh2014nice,rezende2015variational,dinh2016density,kingma2018glow,trippe2018conditional} to define $p(\mathbf{y}|\mathbf{x})$, so that the model is flexible and can represent complex distributions. In training, we use the weak signals $\mathbf{Q}$ to define a constrained space for $\mathbf{y}$ and then optimize the likelihood of all possible $\mathbf{y}$s that are within this constrained space. Therefore, the learned model captures all possible relationships between the input $\mathbf{x}$ and output $\mathbf{y}$. We also develop a learning method for LLF that trains the conditional flow inversely and avoids the complicated min-max optimization~\citep{arachie2021general}. In inference, we use sample-based method~\citep{lu2020structured} to estimate labels for given input.

We apply LLF to three weakly supervised learning problems: weakly supervised classification~\citep{arachie2021general,mazzetto2021semi}, weakly supervised regression, and unpaired point cloud completion~\citep{chen2019unpaired,wu2020multimodal}. These three problems have very different label types and weak signals. Our method outperforms all other state-of-the-art methods on weakly supervised classification and regression, and it can perform comparably to recent methods on unpaired point cloud completion. The experiments show that LLF is versatile and powerful.

\section{Background}
\label{sec:background}

In this section, we introduce weakly supervised learning and conditional normalizing flows.

\subsection{Weakly Supervised Learning.} Given a dataset $\mathcal{D} = \{\mathbf{x}_1,...,\mathbf{x}_N\}$, and weak signals $\mathbf{Q}$, weakly supervised learning finds a model that can predict the unknown label $\mathbf{y}_i$ for each input data $\mathbf{x}_i$. Weak signals are inexact or noisy supervision information that weakly label the dataset~\citep{zhou2018brief}. For different problems, the type and format of weak signals can be different. In weakly supervised classification~\citep{arachie2021general,mazzetto2021semi}, weak signals are noisy labels generated by rule-based labeling methods. In unpaired point cloud completion~\citep{chen2019unpaired,wu2020multimodal}, weak signals are coarse shape and structure information provided by a set of complete point clouds. Detailed illustrations of weakly supervised learning problems are in Section~\ref{sec:case}.

Constraint-based methods~\citep{balsubramani2015scalable,arachie2020constrained,arachie2021general,mazzetto:icml21,mazzetto2021semi}, define a set of constrained functions based on $\mathbf{Q}$ and $\mathbf{x}$. These functions form a space of possible $\mathbf{y}$, and then look for one possible $\mathbf{y}$ within this constrained space. In this work, we follow this idea and use constrained functions to restrict the predicted $\mathbf{y}$. 

\subsection{Conditional Normalizing Flows.} A normalizing flow~\citep{rezende2015variational} is a series of invertible functions $\mathbf{f} = \mathbf{f}_1 \circ \mathbf{f}_2 \circ \dots \circ \mathbf{f}_K$ that transform the probability density of output variables $\mathbf{y}$ to the density of latent variables $\mathbf{z}$. In conditional flows~\citep{trippe2018conditional}, a flow layer function $\mathbf{f}_i$ is also parameterized by the input variables $\mathbf{x}$, i.e., $\mathbf{f}_i = \mathbf{f}_{x, \phi_i}$, where $\phi_i$ is the parameters of $\mathbf{f}_i$. With the change-of-variable formula, the log conditional distribution $\log p(\mathbf{y}|\mathbf{x})$ can be exactly and tractably computed as
\begin{equation}
\label{eq:cflow}
\log p(\mathbf{y}|\mathbf{x}) = \log p_Z(\mathbf{f}_{\mathbf{x}, \phi}(\mathbf{y})) + \sum_{i=1}^{K} \log \left|\det \left(\frac{\partial \mathbf{f}_{\mathbf{x}, \phi_i}}{\partial \mathbf{r}_{i-1}}\right)\right|,
\end{equation}
where $p_Z(\mathbf{z})$ is a tractable base distribution, e.g. Gaussian distribution. The $\frac{\partial \mathbf{f}_{\mathbf{x}, \phi_i}}{\partial \mathbf{r}_{i-1}}$ is the Jacobian matrix of $\mathbf{f}_{\mathbf{x}, \phi_i}$. The $\mathbf{r}_i = \mathbf{f}_{\mathbf{x}, \phi_i}(\mathbf{r}_{i-1})$, $\mathbf{r}_{0} = \mathbf{y}$, and $\mathbf{r}_{K}=\mathbf{z}$.

Normalizing flows are powerful when flow layers are invertible and have tractable Jacobian determinant. This combination enables tractable computation and optimization of exact likelihood. In this paper, we use affine coupling layers~\citep{dinh2014nice,dinh2016density} to form normalizing flows. Affine coupling splits the input to two parts and forces the first part to only depend on the second part, so that the Jacobian is a triangular matrix. For conditional flows, we can define a conditional affine coupling layer as
\begin{align*}
	\mathbf{y}_a, \mathbf{y}_b &= \text{split}(\mathbf{y}), \nonumber\\ 
	\mathbf{z}_b &= \mathbf{s}(\mathbf{x}, \mathbf{y}_a) \odot \mathbf{y}_b  + \mathbf{b}(\mathbf{x}, \mathbf{y}_a), \\
	\quad \mathbf{z} &= \text{concat}(\mathbf{y}_a, \mathbf{z}_b),
\end{align*}
where $\mathbf{s}$ and $\mathbf{b}$ are two neural networks. The $\text{split}()$ function split the input $\mathbf{y}$ to two variabls $\mathbf{y}_a, \mathbf{y}_b$, and the $\text{concat}()$ function concatenates them to one variable.

\section{Proposed Method}
\label{sec:method}

In this section, we introduce the \emph{label learning flows} (LLF) framework for weakly supervised learning. Given $\mathbf{Q}$ and $\mathbf{x}$, we define a set of constraints to restrict the predicted label $\mathbf{y}$. These constraints can be inequalities, formatted as $c(\mathbf{x},\mathbf{y}, \mathbf{Q}) \le b$, or equalities, formatted as $c(\mathbf{x},\mathbf{y}, \mathbf{Q}) = b$. For simplicity, we represent this set of constraints as $\mathbf{C}(\mathbf{x}, \mathbf{y}, \mathbf{Q})$. Let $\Omega$ be the  constrained sample space of $\mathbf{y}$ defined by $\mathbf{C}(\mathbf{x}, \mathbf{y}, \mathbf{Q})$ (when specifying a sample $\mathbf{x}_i$, we use $\Omega_i$ to represent the constrained space of $\mathbf{y}_i$, i.e., the $\Omega_i$ is a specific instance of $\Omega$). 

For each $\mathbf{x}_i$, previous methods~\citep{balsubramani2015scalable,arachie2020constrained,arachie2021general,mazzetto:icml21,mazzetto2021semi} only look for a single best $\mathbf{y}_i$ within $\Omega_i$, resulting in a saddle-point optimization problem:
\begin{equation}
\label{eq:prev_method_obj}
\min_{\theta}\max_{\mathbf{y}\in\Omega} \mathcal{L}(\theta, \mathbf{x}, \mathbf{y}),
\end{equation}
where $\mathcal{L}()$ is the loss function, and $\mathbf{\theta}$ is the model parameter. However, since the constrained space may be loose, there are usually more than one valid labels within $\Omega$. These methods omit this uncertainty, and only learn a deterministic mapping between each $\mathbf{x}_i$ and its optimized label $\mathbf{y}_i$. Besides, optimizing $\mathcal{L}(\mathbf{\theta}, \mathbf{x}, \mathbf{y})$ requires an EM-like algorithm, and optimize $\mathbf{\theta}$ and $\mathbf{y}$ alternatively, which complicates the training process.

To address the above issues, we develop a framework that optimizes the conditional log-likelihood of all possible $\mathbf{y}_i$ within $\Omega_i$, resulting in the objective
\begin{equation}
\label{eq:nolabel_raw}
	\max_{\mathbf{\phi}} \mathbb{E}_{p_{\text{data}}(\mathbf{x})} \mathbb{E}_{\mathbf{y} \sim U(\Omega)}\left[\log p(\mathbf{y}|\mathbf{x}, \phi)\right],
\end{equation}
where $U(\Omega)$ is a distribution of possible $\mathbf{y}$ within $\Omega$, and $p(\mathbf{y}|\mathbf{x}, \phi)$ is a continuous density model of $\mathbf{y}$.

To use the proposed framework, i.e., Eq.~\ref{eq:nolabel_raw}, we need to address two main problems. First, the $p(\mathbf{y}|\mathbf{x})$ should be defined by a flexible model, which has a computable likelihood, and can represent complicated distributions. Second, training the model with Eq.~\ref{eq:nolabel_raw} requires sampling $\mathbf{y}$ within $\Omega$. Using traditional sampling methods, e.g., uniform sampling, to sample $\mathbf{y}$ is inefficient. Due to the high dimensionality of sample space, the rejection rate would be prohibitively high. 

Our method is called label learning flows (LLF), because these two problems can instead be solved with normalizing flows. That is, we use the invertibility of flows and rewrite $\log p(\mathbf{y}|\mathbf{x})$ as
\begin{align}
\label{eq:inverse_flow}
	\log p(\mathbf{y}|\mathbf{x}) &= \log p_Z(\mathbf{f}_{\mathbf{x}, \phi}(\mathbf{y})) + \sum_{i=1}^{K} \log \left|\det \left(\frac{\partial \mathbf{f}_{\mathbf{x}, \phi_i}}{\partial \mathbf{r}_{i-1}}\right)\right| \nonumber \\
	&= \log p_Z(\mathbf{z}) - \sum_{i=1}^{K} \log \left|\det \left(\frac{\partial \mathbf{g}_{\mathbf{x}, \phi_i}}{\partial \mathbf{r}_{i}}\right)\right|,
\end{align}
where $\mathbf{g}_{\mathbf{x}, \phi_i} = f^{-1}_{\mathbf{x}, \phi_i}$ is the inverse flow, and $\mathbf{r}_{i}$ is an intermediate variable output from the $i$-th flow layer.

We first sample $\mathbf{z}$ from $p_{Z}(\mathbf{z})$, and then transform $\mathbf{z}$ samples to $\mathbf{y}$ samples with the inverse flow, i.e., $\mathbf{g}_{\mathbf{x}, \phi}(\mathbf{z})$. We use a set of constraints $\mathbf{C}(\mathbf{x}, \mathbf{g}_{\mathbf{x}, \phi}(\mathbf{z}), \mathbf{Q})$ to restrict the generated $\mathbf{y}$ samples to be within $\Omega$, so that we will be able to efficiently get samples of $\mathbf{y}$ within $\Omega$. Therefore, the Eq.~\ref{eq:nolabel_raw} can be approximately rewritten as a constrained optimization problem
\begin{align}
\label{eq:no_label_final}
	& \max_{\mathbf{\phi}} \mathbb{E}_{p_{\text{data}}(\mathbf{x})}\mathbb{E}_{p_{Z}(\mathbf{z})}\left[ \log p_Z(\mathbf{z}) - \sum_{i=1}^{K} \log \left|\det \left(\frac{\partial \mathbf{g}_{\mathbf{x}, \phi_i}}{\partial \mathbf{r}_{i}}\right)\right|  \right],\nonumber\\
	& \text{s.t.}~~~ \mathbf{C}(\mathbf{x}, \mathbf{g}_{\mathbf{x}, \phi}(\mathbf{z}), \mathbf{Q}).
\end{align}
The Eq.~\ref{eq:no_label_final} is the final LLF framework. In LLF, the problem of sampling $\mathbf{y}$ within $\Omega$ is converted to sampling $\mathbf{z}$, so that can be easily solved. 

For efficient training, this constrained optimization problem can be approximated with the penalty method, resulting in the objective
\begin{align}
\label{eq:nolabel_obj}
	\max_{\mathbf{\phi}} \mathbb{E}_{p_{\text{data}}(\mathbf{x})}\mathbb{E}_{p_{Z}(\mathbf{z})}
	&\left[ \log p_Z(\mathbf{z}) - \sum_{i=1}^{K} \log \left|\det \left(\frac{\partial \mathbf{g}_{\mathbf{x}, \phi_i}}{\partial \mathbf{r}_{i}}\right)\right| \right. \nonumber\\
	& \left. - \mathbf{\lambda}\mathbf{C}_r(\mathbf{x}, \mathbf{g}_{\mathbf{x}, \phi}(\mathbf{z}), \mathbf{Q}) \right],
\end{align}
where $\lambda$ is the penalty coefficient, and $\mathbf{C}_r()$ means we reformulate the constraints to be penalty losses. For example, an inequality constraint will be redefined as a hinge loss.  In training, the inverse flow, i.e., $\mathbf{g}_{\mathbf{x}, \phi}(\mathbf{z})$ estimates $\mathbf{y}$ and computes the likelihood simultaneously, removing the need of EM-like methods and making the training straightforward.

In practice, the expectation with respect to $p_{Z}(\mathbf{z})$ can be approximated with a Monte Carlo estimate with $L_t$ samples. Since we only need to obtain stochastic gradients, we follow previous works~\citep{kingma2013auto} and set $L_t=1$.

Given a trained model and a data point $\mathbf{x}_i$, prediction requires outputting a label $\mathbf{y}_i$ for $\mathbf{x}_i$. We follow~\citep{lu2020structured} and use a sample average, i.e., $\mathbf{y}_i = \sum_{j=1}^{L_p} \mathbf{g}_{\mathbf{x}_i, \phi}(\mathbf{z}_j)$ as the prediction, where $L_p$ is the number of samples used for inference. In our experiments, we found that $L_p=10$ is enough for generating high-quality labels.

\section{Case Studies}
\label{sec:case}

In this section, we illustrate how to use LLF to address weakly supervised learning problems.

\subsection{Weakly Supervised Classification}

We follow previous works~\citep{arachie2021general,mazzetto2021semi} and consider binary classification. For each example, the label $\mathbf{y}$ is a two-dimensional vector within a one-simplex. That is, the $\mathbf{y} \in \mathcal{Y} = \{\mathbf{y} \in [0,1]^2: \sum_{j}\mathbf{y}^{[j]} = 1\}$, where $\mathbf{y}^{[j]}$ is the $j$th dimension of $\mathbf{y}$. Each ground truth label $\hat{\mathbf{y}} \in \{0,1\}^2$ is a two-dimensional one-hot vector. We have $M$ weak labelers, which will generate $M$ weak signals for each data point $\mathbf{x}_i$, i.e., $\mathbf{q}_i = [\mathbf{q}_{i,1},...,\mathbf{q}_{i,M}]$. Each weak signal $\mathbf{q}_{i,m} \in \mathcal{Q} = \{\mathbf{q} \in[0,1]^2: \sum_{j}\mathbf{q}^{[j]}=1\}$ is a soft labeling of the data. In practice, if a weak labeler $m$ fails to label a data point $\mathbf{x}_i$, the $\mathbf{q}_{i,m}$ can be null, i.e., $\mathbf{q}_{i,m} = \emptyset$~\citep{arachie2020constrained}. Following~\citep{arachie2021general}, we assume we have access to error rate bounds of these weak signals $\mathbf{b} = [\mathbf{b}_1,..,\mathbf{b}_M]$. These error rate bounds can be estimated based on empirical data, or set as constants~\citep{arachie2020constrained}. Therefore, the weak signals imply constraints
\begin{align}
\label{eq:weak_classification_constraint}
    &\sum_{\substack{i=1,\\ \mathbf{q}_{i,m} \not= \emptyset}}^N (1-\mathbf{y}_i^{[j]}) \mathbf{q}_{i,m}^{[j]} + \mathbf{y}_i^{[j]} (1-\mathbf{q}_{i,m}^{[j]}) \le N_m \mathbf{b}_m^{[j]} \nonumber\\
    &\forall m \in\{1,...,M\},~~~\forall j \in \{0,1\}
\end{align}
where $N_m$ is the number of data points that are labeled by weak labeler $m$. Eq.~\ref{eq:weak_classification_constraint} roughly restricts the difference between estimated labels and weak signals is bounded by the error rate bound.

This problem can be solved with LLF, i.e., Eq.~\ref{eq:no_label_final}, by defining $\mathbf{C}(\mathbf{x}, \mathbf{g}_{\mathbf{x}, \phi}(\mathbf{z}, \mathbf{Q}))$ to be a combination of weak signal constraints, i.e., Eq.~\ref{eq:weak_classification_constraint}, and simplex constraints, i.e., $\mathbf{y} \in \mathcal{Y}$. The objective function of LLF for weakly supervised classification is
\begin{align}
\label{eq:weak_classification_obj}
\max_{\mathbf{\phi}} & \log p_Z(\mathbf{z}) - \sum_{i=1}^{K} \log \left|\det \left(\frac{\partial \mathbf{g}_{\mathbf{x}, \phi_i}}{\partial \mathbf{r}_{i}}\right)\right| \nonumber\\
&- \lambda_1 \left[\mathbf{g}_{\mathbf{x}, \phi}(\mathbf{z})\right]_{+}^2 - \lambda_2 \left[1-\mathbf{g}_{\mathbf{x}, \phi}(\mathbf{z})\right]_{+}^2 \nonumber\\ 
&- \lambda_3 \left(\sum_{j}\mathbf{g}_{\mathbf{x}, \phi}(\mathbf{z})^{[j]} - 1\right)^2 \nonumber\\
& - \lambda_4 \sum_{j=0}^1 \sum_{m=1}^M\left[\sum_{\substack{i=0 \\ \mathbf{q}_{i,m} \not= \emptyset}}^N (1-\mathbf{g}_{\mathbf{x}, \phi}(\mathbf{z})_i^{[j]}) \mathbf{q}_{i,m}^{[j]} \right. \nonumber\\ 
& \left. + \mathbf{g}_{\mathbf{x}, \phi}(\mathbf{z})_i^{[j]} (1-\mathbf{q}_{i,m}^{[j]}) - N_m \mathbf{b}_m^{[j]} \right]_+^2,
\end{align}
where the second and third rows are the simplex constraints, and the last term is the weak signal constraints reformulated from Eq.~\ref{eq:weak_classification_constraint}. The $[.]_+$ is a hinge function that returns its input if positive and zero otherwise. We omit the expectation terms for simplicity.

\subsection{Weakly Supervised Regression}
For weakly supervised regression, we predict one-dimensional continuous labels $y \in [0,1]$ given input dataset $\mathcal{D} = \{\mathbf{x}_1,...,\mathbf{x}_N\}$ and weak signals $\mathbf{Q}$. We define the weak signals as follows. For the $m$-th feature of input data, we have access to a threshold $\epsilon_m$, which splits $\mathcal{D}$ to two parts, i.e., $\mathcal{D}_{m, 1}, \mathcal{D}_{m, 2}$, such that for each $\mathbf{x}_i \in \mathcal{D}_{m, 1}$, the $\mathbf{x}_{i,m} \ge \epsilon_m$, and for each $\mathbf{x}_j \in \mathcal{D}_{m, 2}$, the $\mathbf{x}_{j,m} < \epsilon_m$. We also have access to estimated values of labels for subsets $\mathcal{D}_{m, 1}$ and $\mathcal{D}_{m, 2}$, i.e., $b_{m,1}$ and $b_{m,2}$. This design of weak signals tries to mimic that in practical scenarios, human experts can design rule-based methods for predicting labels for given data. For example, marketing experts can predict the prices of houses based on their size. For houses whose size is greater than a threshold, an experienced expert would know an estimate of their average price. Assuming that we have $M$ rule-based weak signals, the constraints can be defined as follows:
\begin{align}
\label{eq:regression_constraints}
    &\frac{1}{|\mathcal{D}_{m,1}|}\sum_{i \in \mathcal{D}_{m,1}} y_i = b_{m,1},~~~\frac{1}{|\mathcal{D}_{m,2}|}\sum_{j \in \mathcal{D}_{m,2}} y_j = b_{m,2},\nonumber\\
    & \forall m \in \{1,...,M\}.
\end{align}

Plugging in Eq.~\ref{eq:regression_constraints} to Eq.~\ref{eq:nolabel_obj}, we have 
\begin{align}
\max_{\mathbf{\phi}}&\log p_Z(z) - \sum_{i=1}^{K} \log \left|\det \left(\frac{\partial \mathbf{g}_{\mathbf{x}, \phi_i}}{\partial r_{i}}\right)\right| \nonumber\\
& - \lambda_1 \left[\mathbf{g}_{\mathbf{x}, \phi}(z)\right]_{+}^2 - \lambda_2 \left[1-\mathbf{g}_{\mathbf{x}, \phi}(z)\right]_{+}^2 \nonumber \\
& - \lambda_3 \sum_{m=1}^M\left( \left(\frac{1}{|\mathcal{D}_{m,1}|}\sum_{i \in \mathcal{D}_{m,1}} \mathbf{g}_{\mathbf{x}, \phi}(z)_i - b_{m,1}\right)^2 \right.\nonumber\\
& \left. + \left(\frac{1}{|\mathcal{D}_{m,2}|}\sum_{j \in \mathcal{D}_{m,2}} \mathbf{g}_{\mathbf{x}, \phi}(z)_j - b_{m,2}\right)^2 \right), \nonumber\\
\end{align}
where the first two constraints restrict $y \in [0,1]$. The last two rows are the weak signal constraints reformulated from Eq.~\ref{eq:regression_constraints}.

\subsection{Unpaired Point Cloud Completion}

Unpaired point cloud completion~\citep{chen2019unpaired,wu2020multimodal} is a practical problem in 3D scanning. Given a set of partial point clouds $\mathcal{X}_p = \{\mathbf{x}_1^{(p)},...,\mathbf{x}_N^{(p)}\}$, and a set of complete point clouds $\mathcal{X}_c = \{\mathbf{x}_1^{(c)},...,\mathbf{x}_N^{(c)}\}$, we want to restore each $\mathbf{x}_i^{(p)} \in \mathcal{X}_p$ by generating its corresponding complete and clean point cloud $\mathbf{y}_i \in \mathcal{Y}$. Each point cloud is a set of points, e.g., $\mathbf{x}_i = \{\mathbf{x}_{i,1},...,\mathbf{x}_{i,T}\}$, where $\mathbf{x}_{i,t} \in \mathcal{R}^3$ is a 3D point, and the counts $T$ represent the number of points in a point cloud. Note that the point clouds in $\mathcal{X}_p$ and $\mathcal{X}_c$ are \emph{unpaired}, so directly modeling their relationship with supervised learning is impossible. 

However, this problem can be interpreted as an inexact supervised learning problem~\citep{zhou2018brief}. That is, the weak supervision information, e.g., the shape and structure of 3D objects, is given by the referred complete point clouds $\mathcal{X}_c$. To capture this information, \citeauthor{chen2019unpaired} and \citeauthor{wu2020multimodal} propose to use adversarial learning and  train a discriminator of least square GAN~\citep{mao2017least} with the referred complete point clouds. This discriminator $D()$ provides a score within $[0, 1]$ to each generated complete point cloud, indicating its quality and fidelity, i.e., a higher score indicates a more realistic complete point cloud. This weak signal provided by $D()$ can be written as an equality constraint
\begin{align}
    \label{eq:pc_gan_constraint}
    D(\mathbf{y}_i) = 1, \quad\quad i \in \{1, ..., N\}.
\end{align}
 
 The conditional distribution $p(\mathbf{y}|\mathbf{x}_p)$ is an exchangeable distribution. We follow previous works~\citep{yang2019pointflow,klokov2020discrete} and use De Finetti’s representation theorem and variational inference to compute its lower bound as the objective.
\vspace{-5mm}
\begin{align}
\log p(\mathbf{y} | \mathbf{x}_p) &\ge  \mathbb{E}_{q(\mathbf{u}|\mathbf{x}_p)} \left[ \sum_{i=1}^{T_c} \log p(\mathbf{y}_{i} | \mathbf{u}, \mathbf{x}_p)\right] \nonumber\\ 
&- \text{KL}(q(\mathbf{u}|\mathbf{x}_p) || p(\mathbf{u})),
\end{align}
where $T_c$ is the number of points of a complete shape. The $q(\mathbf{u}|\mathbf{x}_p)$ is a variational distribution of latent variable $\mathbf{u}$. In practice, it can be represented by an encoder, and uses the reparameterization trick~\citep{kingma2013auto} to sample $\mathbf{u}$. The $p(\mathbf{u})$ is a standard Gaussian prior. The $p(\mathbf{y}_i|\mathbf{u},\mathbf{x}_p)$ is defined by a conditional flow. The final objective function is 
\begin{align}
\label{eq:point_cloud_obj}
\max_{\phi}& \mathbb{E}_{q(\mathbf{u}|\mathbf{x}_p)} \left[ \sum_{t=1}^{T_c} \log p_Z(\mathbf{z_t}) - \sum_{i=1}^{K} \log \left|\det \left(\frac{\partial \mathbf{g}_{\mathbf{u}, \mathbf{x}_p, \phi_i}}{\partial \mathbf{r}_{t,i}}\right)\right|\right] \nonumber\\ 
& - \text{KL}(q(\mathbf{u}|\mathbf{x}_p) || p(\mathbf{u})) \nonumber\\
& - \mathbb{E}_{q(\mathbf{u}|\mathbf{x}_p)} \left[ \lambda_1 (D(\mathbf{g}_{\mathbf{u}, \mathbf{x}_p,\phi}(\mathbf{z})) -1)^2 \right. \nonumber\\
& \left. + \lambda_2 d_{H}(\mathbf{g}_{\mathbf{u}, \mathbf{x}_p, \phi}(z), \mathbf{x}_p) \right],
\end{align}
where the $d_{H}()$ represents the Haudorsff distance~\citep{chen2019unpaired}, which measures the distance between a generated complete point cloud and its corresponding input partial point cloud. The third term is reformatted from Eq.~\ref{eq:pc_gan_constraint}. For clarity, we use $\mathbf{z}_t$ and $\mathbf{r}_t$ to represent variables of the $t$-th point in a point cloud, and $\mathbf{g}_{\mathbf{u}, \mathbf{x}_p,\phi}(\mathbf{z})$ to represent a generated point cloud. Detailed derivations of Eq.~\ref{eq:point_cloud_obj} are in appendix.

Training with Eq.~\ref{eq:point_cloud_obj} is different from the previous settings, because we also need to train the discriminator of the GAN. The objective for $D()$ is
\begin{align}
\label{eq:gan_dis_obj}
    \min_{D} & \mathbb{E}_{ p_{\text{data}}(\mathbf{x}_c)}\left[ (D(\mathbf{x}_c)-1)^2\right] \nonumber\\
    & + \mathbb{E}_{ p_{\text{data}}(\mathbf{x}_p),p_{Z}(\mathbf{z}),q(\mathbf{u}|\mathbf{x}_p)}
     \left[D(\mathbf{g}_{\mathbf{x}_p,\mathbf{u},\phi}(\mathbf{z}))^2 \right].
\end{align}

The training process is similar to traditional GAN training. The inverse flow $\mathbf{g}_{\mathbf{u},\mathbf{x}_p, \phi}$ can be roughly seen as the generator. In training, we train the flow to optimize Eq.~\ref{eq:point_cloud_obj} and the discriminator to optimize Eq.~\ref{eq:gan_dis_obj}, alternatively.

\section{Related Work}
\label{sec:relate}

In this section, we introduce the research that most related to our work.

\subsection{Weakly Supervised Learning.} Our method is in the line of constraint-based weakly supervised learning~\citep{balsubramani2015scalable,arachie2020constrained,arachie2021general,mazzetto:icml21,mazzetto2021semi}, which constrains the label space of the predicted labels using weak supervision and estimated errors. Previous methods are deterministic and developed for classification tasks specifically. They estimate one possible $\mathbf{y}$ within the constrained space $\Omega$. In contrast to these methods, LLF learns a probabilistic model, i.e., conditional flow, to represent the relationship between $\mathbf{x}$ and $\mathbf{y}$. In training, it optimizes the likelihoods of all possible $\mathbf{y}$s within $\Omega$. For weakly supervised classification, LLF uses the same strategy as adversarial label learning (ALL)~\citep{arachie2021general} to define constraint functions based on weak signals. ALL then uses a min-max optimization to learn the model parameters and estimate $\mathbf{y}$ alternatively. Unlike ALL, LLF learns the model parameters and output $\mathbf{y}$ simultaneously, and does not need a min-max optimization. Besides, LLF is a general framework and can be applied to other weakly supervised learning problems. 

In another line of research, non-constraint based weak supervision methods~\citep{ratner2016data,ratner2019training,fu2020fast,shin2021universalizing,kuang2022firebolt} typically assume a joint distribution for the weak signals and the ground-truth labels. These methods use graphical models to estimate labels while accounting for dependencies among weak signals. Recently, WeaSEL~\citep{ruhling2021end} reparameterizes graphical models with an encoder network. Recently, WeaNF~\citep{2022WeaNF} uses normalizing flows to model labeling functions that output weak labels. Unlike these methods, we use a flow network to model dependency between $\mathbf{x}$ and $\mathbf{y}$, but does not consider relationships among weak signals.

Besides, there are other weakly supervised learning methods. Active WeaSuL~\citep{biegel2021active} incorporates active learning into weakly supervised learning. Losses over Labels~\citep{sam2023losses} optimizes loss functions that are derived from weak labelers. Some recent methods~\citep{yu2020fine,karamanolakis2021self} develop self-training frameworks to train neural networks with weak supervision. \citeauthor{zhang2021wrench} develop a benchmark for weak supervision.

\subsection{Normalizing Flows.} Normalizing flows~\citep{dinh2014nice,rezende2015variational,dinh2016density,kingma2018glow} have gained recent attention because of their advantages of exact latent variable inference and log-likelihood evaluation. Specifically, conditional normalizing flows have been widely applied to many supervised learning problems ~\citep{trippe2018conditional,lu2020structured,lugmayr2020srflow,pumarola2020c} and semi-supervised classification~\citep{atanov2019semi,izmailov2020semi}. However, normalizing flows have not previously been applied to weakly supervised learning problems.

Our inverse training method for LLF is similar to injective flows~\citep{kumar2020regularized}. Injective flows are used to model unconditional datasets. They use an encoder network to map the input data $\mathbf{x}$ to latent code $\mathbf{z}$, and they use an inverse flow to map $\mathbf{z}$ back to $\mathbf{x}$, resulting in an autoencoder architecture. Different from injective flow, LLF directly samples $\mathbf{z}$ from a prior distribution and uses a conditional flow to map $\mathbf{z}$ back to $\mathbf{y}$ conditioned on $\mathbf{x}$. We use constraint functions to restrict $\mathbf{y}$ to be valid, so that does not need an encoder network.

\subsection{Point Cloud Modeling.} Recently, \cite{yang2019pointflow} and \cite{tran2019discrete} combine normalizing flows with variational autoencoders~\citep{kingma2013auto} and developed continuous and discrete normalizing flows for point clouds. The basic idea of point normalizing flows is to use a conditional flow to model each point in a point cloud. The conditional flow is conditioned on a latent variable generated by an encoder. To guarantee exchangeability, the encoder uses a PointNet~\citep{qi2017pointnet} to extract features from input point clouds.

The unpaired point cloud completion problem is defined by \citep{chen2019unpaired}. They develop pcl2pcl---a GAN~\citep{goodfellow2014generative} based model---to solve it. Their method is two-staged. In the first stage, it trains autoencoders to map partial and complete point clouds to their latent space. In the second stage, a GAN is used to transform the latent features of partial point clouds to latent features of complete point clouds. In their follow-up paper~\citep{wu2020multimodal}, they develop a variant of pcl2pcl, called multi-modal pcl2pcl (mm-pcl2pcl), which incorporates random noise to the generative process, so that can capture the uncertainty in reasoning.

Aside from pcl2pcl, some other GAN-based methods developed recently for this problem. Cycle4Completion~\citep{wen2021cycle4completion} uses inverse cycle transformations to improve completion
accuracy of 3D shapes. Shape-inversion~\citep{zhang2021unsupervised} do shape completion by using a GAN pre-trained on complete shapes. Given a partial input, it looks for the best latent code that can reconstruct the input shape. KT-Net~\citep{cao2023kt} develops a Teacher-Assistant-Student framework to transfer knowledge from complete shape domain to incomplete shape domain. 

When applying LLF to this problem, LLF has a similar framework to VAE-GAN~\citep{larsen2016autoencoding}. The main differences are that LLF models a conditional distribution of points, and its encoder is a point normalizing flow. Besides, different from pcl2pcl, LLF can be trained end-to-end.

\section{Empirical Study}
\label{sec:experiment}

In this section, we evaluate LLF on the three weakly supervised learning problems.

%\subsection{Architecture and Setup}
\textbf{Model architecture.} For weakly supervised classification and unpaired point cloud completion, the labels $\mathbf{y}$ are multi-dimensional variables. We follow \citep{klokov2020discrete} and use flows with only conditional coupling layers. We use the same method as \citep{klokov2020discrete} to define the conditional affine layer. Each flow model contains $8$ flow steps. For unpaired point cloud completion, each flow step has $3$ coupling layers. For weakly supervised classification, each flow step has $2$ coupling layers. For weakly supervised regression, since $y$ is a scalar, we use simple conditional affine transformation as flow layer. The flow for this problem contains $8$ conditional affine transformations.

For the unpaired point cloud completion task, we need to also use an encoder network, i.e., $q(\mathbf{u}|\mathbf{x}_p)$ and a discriminator $D()$. We follow \citep{klokov2020discrete,wu2020multimodal} and use PointNet~\citep{qi2017pointnet} in these two networks to extract features for point clouds.

\textbf{Experiment setup.} In weakly supervised classification and regression experiments, we assume that the ground truth labels are inaccessible, so tuning hyper-parameters for models is impossible. We use default settings for all hyper-parameters of LLF, e.g., $\lambda$s and learning rates. We fix $\lambda = 10$ and use Adam~\citep{kingma2014adam} with default settings. Following previous works~\citep{arachie2021general,arachie2020constrained}, we use full gradient optimization to train the models. For fair comparison, 
we run each experiment $5$ times with different random seeds. For experiments with unpaired point cloud completion, we tune the hyper-parameters using validation sets. We use stochastic optimization with Adam to train the models. More details about experiments are in the appendix.

% $\{0,10,100,123,1234\}$.

% i.e., $\eta = 0.001$, $\beta_1=0.9$ and $\beta_2=0.999$. We use an exponential learning rate scheduler with a decreasing rate of $0.996$ to guarantee convergence. We track the decrease of loss and when the decrease is small enough, the training stops. Following previous works~\cite{arachie2021general,arachie2020constrained}, we use full gradient optimization to train the models. For fair comparison, 
% we run each experiment $5$ times with different random seeds $\{0,10,100,123,1234\}$. We split each dataset to training, simulation, and test sets. We follow \cite{arachie2021general,arachie2020constrained} and create weak signals with randomly chosen features, and estimate error bounds and thresholds on simulation set.

% with an initial learning rate $\eta = 0.0001$ and default $\beta$s. The best coefficients for the constraints in Equation~\ref{eq:point_cloud_obj} are $\lambda_1=10, \lambda_2=100$. We use stochastic optimization to train the models, and the batch size is $32$. Each model is trained for at most $2000$ epochs. 

\subsection{Weakly Supervised Classification}

\textbf{Datasets.} We follow \cite{arachie2020constrained,arachie2021general} and conduct experiments on $12$ datasets. Specifically, the Breast Cancer, OBS Network, Cardiotocography, Clave Direction, Credit Card, Statlog Satellite, Phishing Websites, Wine Quality are tabular datasets from the UCI repository~\citep{Dua2019}. The Fashion-MNIST~\citep{xiao2017fashion} is an image set with 10 classes of clothing types. We choose $3$ pairs of classes, i.e.,
dresses/sneakers (DvK), sandals/ankle boots (SvA), and coats/bags (CvB), to conduct binary classification.  We follow \citep{arachie2021general} and create $3$ synthetic weak signals for each dataset. Each dataset is split to training set, simulation set and test set. The error rate bounds are estimated based on the simulation set. The IMDB~\citep{maas2011learning}, SST~\citep{socher2013recursive} and YELP are real text datasets. We follow \citep{arachie2020constrained} and use keyword-based weak supervision. Each dataset has more than $10$ weak signals. The error rate bounds are set as $0.01$. 

\begin{table*}[!htp]
\vspace{-0.2cm}
	\begin{center}
		\caption{Test set accuracy (in percentage) on tabular and image datasets. We report the mean accuracy of $5$ experiments, and the subscripts are standard deviation.}
		\label{tab:classification_synthetic}
		\footnotesize
		\begin{tabular}{l| l l l l l l| l}
			\toprule
			~ & LLF & ALL & PGMV & ACML & GE & AVG & SL \\
			\midrule
			Fashion MNIST (DvK) & $\mathbf{100_{0.0}}$ & $99.5_{0.0}$ & $50.15_{0.0}$ & $75.65_{0.0}$ & $97.9_{0.0}$ & $83.5_{0.0}$ & $1.0_{0.0}$  \\
			Fashion MNIST (SvA) & $\mathbf{94.4_{ 0.1}}$ & $90.8_{0.0}$ & $56.15_{0.0}$ & $71.45_{0.0}$ & $50.1_{0.0}$ & $79.1_{0.0}$ & $97.2_{0.0}$ \\
			Fashion MNIST (CvB) & $\mathbf{91.6_{3.8}}$ & $80.5_{0.0}$ & $56.45_{0.0}$ & $68.75_{0.0}$ & $50.1_{0.0}$ & $74.0_{0.0}$ &$98.8_{0.0}$ \\
			Breast Cancer & $\mathbf{96.8_{0.8}}$ & $93.7_{1.9}$ & $84.10_{2.0}$ & $91.69_{2.4}$ & $93.3_{1.6}$ & $91.1_{2.3}$ & $97.3_{0.7}$ \\
			OBS Network & $68.4_{0.6}$ & $69.1_{1.1}$ & $\mathbf{72.55_{1.7}}$ & $71.71_{1.9}$ & $67.6_{1.0}$ & $70.9_{2.4}$ & $70.4_{3.2}$\\
			Cardiotocography & $93.1_{1.0}$ & $79.5_{1.1}$ & $93.28_{2.2}$ & $\mathbf{94.05_{0.6}}$ & $66.3_{6.1}$ & $90.2_{4.7}$ & $94.1_{0.8}$\\
			Clave Direction & $\mathbf{85.8_{1.7}}$ & $75.0_{1.3}$ & $64.66_{0.5}$ & $70.72_{0.35}$ & $75.6_{2.8}$ & $70.7_{0.3}$ & $96.3_{0.1}$\\
			Credit Card & $\mathbf{68.0_{2.2}}$ & $67.8_{2.1}$ & $57.63_{1.1}$ & $62.38_{3.2}$ & $49.2_{8.8}$ & $60.2_{1.0}$ & $71.7_{3.1}$\\
			Statlog Satellite & $\mathbf{99.7_{0.2}}$ & $95.9_{0.8}$ & $66.22_{0.8}$ & $88.28_{1.1}$ & $98.7_{1.2}$ & $91.5_{1.1}$ & $99.9_{0.1}$\\
			Phishing Websites & $\mathbf{90.6_{0.3}}$ & $89.6_{0.5}$ & $75.71_{3.9}$ & $84.72_{0.2}$ &  $87.0_{0.9}$ & $84.8_{0.2}$ & $92.9_{0.1}$\\
			Wine Quality & $\mathbf{64.7_{1.7}}$ & $62.3_{0.0}$ & $59.61_{0.0}$ & $59.61_{0.0}$ & $44.5_{1.4}$ & $55.5_{0.0}$ & $68.5_{0.0}$\\
			\bottomrule
		\end{tabular}
	\end{center}
	% \vspace{-0.5cm}
\end{table*}

\begin{table*}[!htp]
\vspace{-0.3cm}
	\begin{center}
		\caption{Test set accuracy (in percentage) on real text datasets.}
		\label{tab:results_real}
		\footnotesize
		\begin{tabular}{l| l l l l l l l| l}
			\toprule
			~ & LLF & LLF-TS & CLL & MMCE & DP & MV & MeTaL & SL \\
			\midrule
			SST & $\mathbf{74.6_{0.3}}$ & $72.9_{0.4}$ & $72.9_{0.1}$ & $72.7$ & $72.0_{0.1}$ & $72.0_{0.1}$ & $72.8_{0.1}$ & $79.2_{0.1}$ \\
			IMDB & $\mathbf{75.2_{0.1}}$ & $75.0_{0.6}$ & $74.0_{0.5}$ & $55.1$ & $62.3_{0.7}$ & $72.4_{0.4}$ & $74.2_{0.4}$ & $82.0_{0.3}$ \\
			YELP & $81.1_{0.1}$ & $78.6_{0.2}$ & $\mathbf{84.0_{0.1}}$ & $68.0$ & $76.0_{ 0.5}$ & $79.8_{0.7}$ & $78.0_{0.2}$ & $87.9_{0.1}$ \\
			\bottomrule
		\end{tabular}
	\end{center}
	% \vspace{-0.4cm}
\end{table*}

% Specifically, the Breast Cancer, OBS Network, Cardiotocography, Clave Direction, Credit Card, Statlog Satellite, Phishing Websites, Wine Quality are tabular datasets from the UCI repository~\cite{Dua2019}. The Fashion-MNIST~\cite{xiao2017fashion} is an image set with 10 classes of clothing types. We choose $3$ pairs of classes, i.e.,
% dresses/sneakers (DvK), sandals/ankle boots (SvA), and coats/bags (CvB), to conduct binary classification. We follow \cite{arachie2021general} and create $3$ synthetic weak signals for each dataset. Each dataset is split to training set, simulation set and test set. The error rate bounds are estimated based on the simulation set. The IMDB~\cite{maas2011learning}, SST~\cite{socher2013recursive} and YELP are real text datasets. We follow \cite{arachie2020constrained} and use keyword-based weak supervision. Each dataset has more than $10$ weak signals. The error rate bounds are set as $0.01$. For experiments on tabular datasets, we set the maximum epochs to $2000$. For experiments on real text datasets, we set the maximum epochs to $500$. 

\textbf{Baselines.} We compare our method with state-of-the-art methods for weakly supervised classification. For the experiments on tabular datasets and image sets, we use ALL~\citep{arachie2021general}, PGMV~\citep{mazzetto2021semi}, ACML~\citep{mazzetto:icml21}, generalized expectation (GE) \citep{druck2008learning,mann2010generalized} and averaging of weak signals (AVG). For experiments on text datasets, we use CLL~\citep{arachie2020constrained}, Snorkel MeTaL~\citep{ratner2019training}, Data Programming (DP)~\citep{ratner2016data},  regularized minimax conditional entropy for crowdsourcing (MMCE)~\citep{zhou2015regularized}, and majority-vote (MV). Note that some baselines, e.g., DP, CLL, used on text datasets are two-stage method. That is, they first predict labels for data points, and then use estimated labels to train downstream classifiers. For better comparison, we develop a two-stage variant of LLF, i.e., LLF-TS, which first infers the labels, and then train a classifier with these inferred labels as a final predictor. We also show supervised learning (SL) results for reference.

\textbf{Results.} We report the mean and standard deviation of accuracy (in percentage) on test sets in Table~\ref{tab:classification_synthetic} and Table~\ref{tab:results_real}. For experiments on tabular and image datasets, LLF outperforms other baselines on $9/11$ datasets. On some datasets, LLF can perform as well as supervised learning methods. For experiments on text datasets, LLF outperforms other baselines on $2/3$ datasets. LLF-TS performs slightly worse than LLF, we feel that one possible reason is LLF is a probabilistic model, which uses the average of samples as estimate labels, so that can slightly smooth out anomalous values, and improve predictions. These results prove that LLF is powerful and effective. In our experiments, we also found that the performance of LLF is impacted by different initialization of weights. This is why LLF has relatively larger variance on some datasets. 
\begin{figure}[!htp]
	\centering
   \includegraphics[width=1\linewidth]{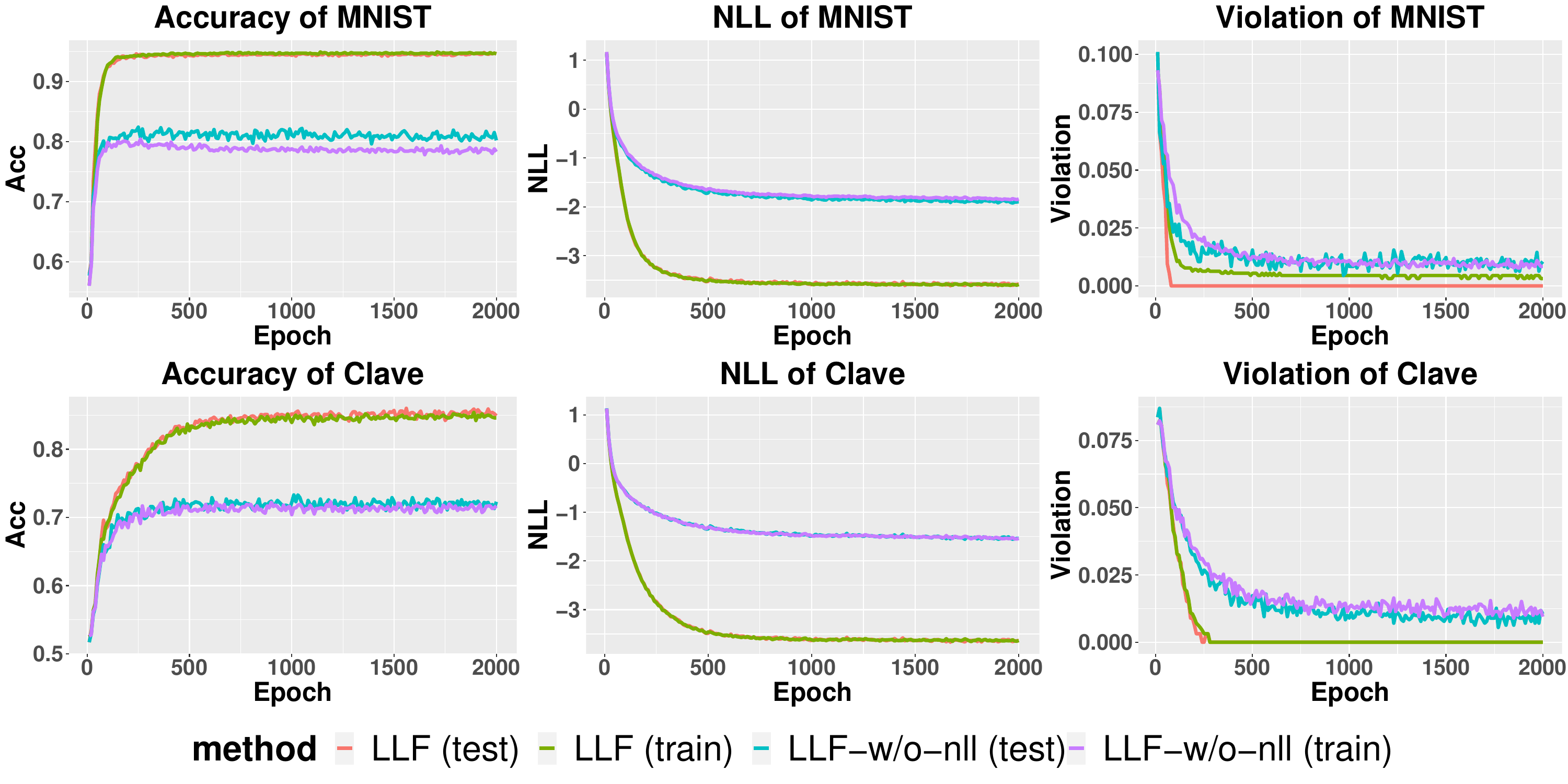}
\caption{Evolution of accuracy, likelihood and violation of weak signal constraints. Training with likelihood makes LLF accumulate more probability mass to the constrained space, so that the generated $\mathbf{y}$ are more likely to be within $\Omega$, and the predictions are more accurate.}
\label{fig:evolution}
	\vspace{-0.1in}
\end{figure}

% \begin{wrapfigure}[16]{r}{0.7\textwidth}\vspace{-0.5cm}
% 	\centering
%    \includegraphics[width=1\linewidth]{2022_neurips_llf/figures/evolution.pdf}
% \caption{Evolution of accuracy, likelihood and violation of weak signal constraints.}
% \label{fig:evolution}
% \end{wrapfigure}

\textbf{Ablation Study.} We can directly train the model using only the constraints as the objective function. In our experiments, we found that training LLF without likelihood (LLF-w/o-nll) will still work. However, the model performs worse than training with likelihood, i.e., Figure~\ref{fig:evolution} We believe that this is because the likelihood helps accumulate more probability mass to the constrained space $\Omega$, so the model will more likely generate $\mathbf{y}$ samples within $\Omega$, and the predictions are more accurate.

\subsection{Weakly Supervised Regression}

\begin{table*}[!htp]
\vspace{-0.2cm}
	\begin{center}
		\caption{Test set RMSE of different methods. The numbers in brackets indicate the label's range.}
		\label{tab:results_regression}
		\footnotesize
		\begin{tabular}{l | l l l| l}
			\toprule
			~ & LLF & LLF-w/o-nll & AVG & SL \\
			\midrule
			Air Quality ($0.1847 \sim
2.231$)& $\mathbf{0.211_{0.009}}$ & $0.266_{0.004}$ & $0.373_{0.005}$ & $0.123_{0.002}$ \\
			Temperature Forecast ($17.4 \sim 38.9$) & $\mathbf{2.552_{0.050}}$ & $2.656_{0.055}$ & $2.827_{0.027}$ & $1.465_{0.031}$ \\
			Bike Sharing ($1 \sim 999$) & $\mathbf{157.348_{0.541}}$ & $162.697_{1.585}$ & $171.338_{1.300}$ & $141.920_{ 1.280}$ \\
			\bottomrule
		\end{tabular}
	\end{center}
	% \vspace{-0.1in}
\end{table*}
\begin{table*}[!htp]
\vspace{-0.2cm}
	\begin{center}
		\caption{Evaluation results on three classes of PartNet. LLF performs comparable to baselines. }
		\label{tab:result_pc}
		\footnotesize
		\begin{tabular}{ l | l  l  l | l l l| lll}
			\toprule
			\textbf{PartNet} & \multicolumn{3}{c|}{\textbf{Chair}} & \multicolumn{3}{c|}{\textbf{Lamp}} & \multicolumn{3}{c}{\textbf{Table}} \\
			\toprule
			~ & MMD$\downarrow$ & TMD$\uparrow$ & UHD$\downarrow$ & MMD$\downarrow$ & TMD$\uparrow$ & UHD$\downarrow$ & MMD$\downarrow$ & TMD$\uparrow$ & UHD$\downarrow$ \\
			\midrule
			LLF & $1.72$ & $0.63$ & $5.74$ & $2.11$ & $0.57$ & $4.71$ & $1.57$ & $0.55$ & $5.42$ \\
			LLF-w/o-nll & $1.79$ & $0.47$ & $5.49$ & $2.21$ & $0.41$ & $4.61$ & $1.57$ & $0.43$ & $5.13$ \\
			pcl2pcl & $1.90$ & $0.00$ & $4.88$ & $2.50$ & $0.00$ & $4.64$ & $1.90$ & $0.00$ & $4.78$ \\ 
			mm-pcl2pcl & $\mathbf{1.52}$ & $\mathbf{2.75}$ & $6.89$ & $\mathbf{1.97}$ & $\mathbf{3.31}$ & $5.72$ & $\mathbf{1.46}$ & $\mathbf{3.30}$ & $5.56$ \\
			mm-pcl2pcl-im & $1.90$ & $1.01$ & $6.65$ & $2.55$ & $0.56$ & $5.40$ & $1.54$ & $0.51$ & $5.38$\\
                shape-inversion & $2.07$ & $0.51$ & $\mathbf{4.59}$ & $2.23$ & $0.44$ & $\mathbf{3.87}$ & $1.97$ & $0.51$ & $\mathbf{4.35}$\\
                KT-net & $1.97$ & $0.00$ & $5.00$ & $3.22$ & $0.00$ & $4.80$ & $2.16$ & $0.00$ & $5.07$\\
			\bottomrule
		\end{tabular}
	\end{center}
\end{table*}
\textbf{Datasets.} We use $3$ tabular datasets from the UCI repository~\citep{Dua2019}: Air Quality, Temperature Forecast, and Bike Sharing dataset. For each dataset, we randomly choose $5$ features to develop the rule based weak signals. We split each dataset to training, simulation, and test sets. The simulation set is then used to compute the threshold $\epsilon$s, and the estimated label values $b$s. Since we do not have human experts to estimate these values, we use the mean value of a feature as its threshold, i.e., $\epsilon_m = \frac{1}{|\mathcal{D}_{\text{valid}}|} \sum_{i \in \mathcal{D}_{\text{valid}}} \mathbf{x}_i[m]$. We then compute the estimated label values $b_{m,1}$ and $b_{m,2}$ based on labels in the simulation set. Note that the labels in the simulation set are only used for generating weak signals, simulating human expertise. In training, we still assume that we do not have access to labels. We normalize the original label to within $[0,1]$ in training, and recover the predicted label to original value in prediction.

\textbf{Baselines.} To the best of our knowledge, there are no methods specifically designed for weakly supervised regression of this form. We use average of weak signals (AVG) and LFF-w/o-nll as baselines. We also report the supervised learning results for reference.

\textbf{Results.} We use root square mean error (RSME) as metric and the results are in Table~\ref{tab:results_regression}. In general, LLF can predict reasonable labels. Its results are much better than AVG or any of the weak signals alone. Similar to the classification results, training LLF without using likelihood will reduce its performance.

\subsection{Unpaired Point Cloud Completion}

\textbf{Datasets.} We use the Partnet~\citep{mo2019partnet} dataset in our experiments. We follow \citep{wu2020multimodal} and conduct experiments on the $3$ largest classes of PartNet: Table, Chair, and Lamp. We treat each class as a dataset, split to training, validation, and test sets based on official splits of PartNet. For each point cloud, we remove points of randomly selected parts to create a partial point cloud. We follow~\citep{chen2019unpaired,wu2020multimodal} and let the partial point clouds have $1024$ points, and the complete point clouds have $2048$ points. 

% We let the latent variable $\mathbf{u}$ of the VAE to be a $128$-dimensional vector.

\textbf{Metrics.} We follow \citep{wu2020multimodal} and use minimal matching distance (MMD)~\citep{achlioptas2018learning}, total mutual difference (TMD), and unidirectional Hausdorff distance (UHD) as metrics. MMD measures the quality of generated complete shapes. A lower MMD is better. TMD measures the diversity of samples. A higher TMD is better. UHD  measures the fidelity of samples. A lower UHD is better. 

\textbf{Baselines.} We compare our method with pcl2pcl~\citep{chen2019unpaired}, mm-pcl2pcl~\citep{wu2020multimodal}, mm-pcl2pcl-im, shape-inversion~\citep{zhang2021unsupervised}, KT-Net~\citep{cao2023kt} and LLF-w/o-nll. Mm-pcl2pcl-im is a variant of mm-pcl2pcl, which jointly trains the encoder of modeling multi-modality and the GAN.

\begin{figure}[!htp]

	\centering
   \includegraphics[width=1.0\linewidth]{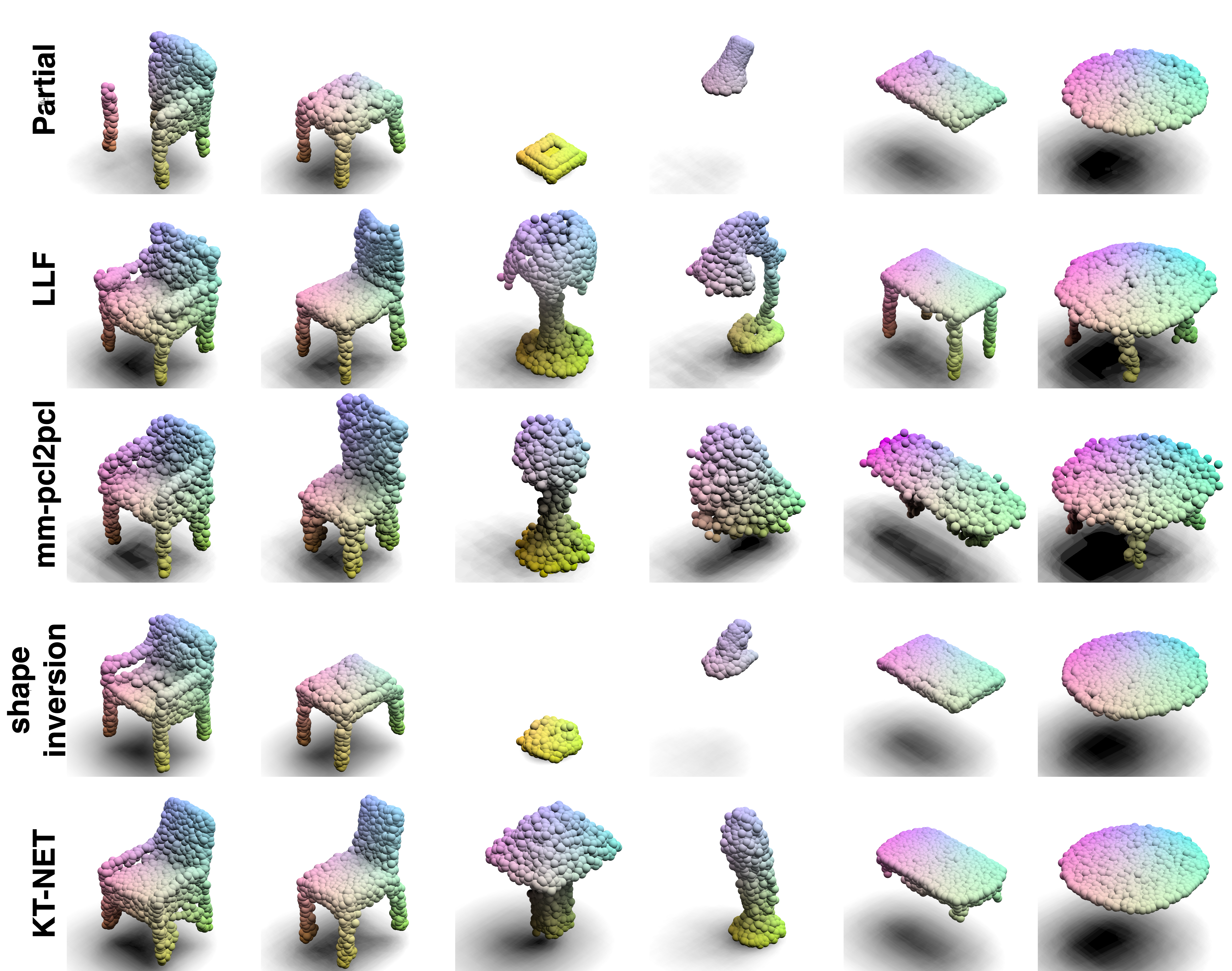}
\caption{Random sample point clouds generated by different methods. The point clouds generated by LLF are as realistic as mm-pc2pc. Mm-pc2pc has a higher diversity in samples. However, sometimes it may generate unreasonable or invalid shapes. Shape-inversion and KT-net cannot generate qualified complete shapes, when the input has limited information, e.g., the input is only a lampshade.}
\label{fig:pc_sample1}
\end{figure}

% \begin{wrapfigure}[13]{r}{0.6\textwidth}
% \vspace{-0.6cm}
% 	\centering
%    \includegraphics[width=1.0\linewidth]{2022_neurips_llf/figures/pc1.pdf}
% \caption{Comparison of random sample point clouds.}
% \label{fig:pc_sample1}
% \end{wrapfigure}

\textbf{Results.} We list the test set results in Table~\ref{tab:result_pc}. In general, pcl2pcl has good fidelity, because it is a discriminative model, and it will only predict one certain sample for each input. This is also why pcl2pcl has the worse diversity as measured by TMD. Mm-pcl2pcl has the best diversity. However, as shown in Figure~\ref{fig:pc_sample1}, some samples generated by mm-pcl2pcl are invalid, i.e., they are totally different from the input partial point clouds. Therefore, mm-pcl2pcl has the worse fidelity, i.e., highest UHD. Mm-pcl2pcl requires a two-stage training. Its end-to-end variant, i.e., mm-pcl2pcl-im performs worse than LLF. Shape-inversion and KT-net have the best UHD, but worst MMD. In our experiments, we notice that these two methods cannot generate valid complete shapes, when the input has only limited geometric and semantic information. For example, in Figure~\ref{fig:pc_sample1}, when the input is only a lampshade, or a tabletop, these two methods cannot recover the complete shapes. Besides, shape-inversion and KT-net have other flaws. The inference of shape-inversion is slow, i.e., it is around $200$ times slower than LLF. The training of KT-net is unstable, and requires fine-tuning hyper-parameters. 

LLF has the second best MMD, i.e., slightly worse than mm-pcl2pcl, but better than other baselines, indicating that it can generate high-quality complete shapes. LLF is better than mm-pcl2pcl on UHD, because most samples generated by LLF are more realistic. It can also generate multi-modal samples, so has better TMD than pcl2pcl and KT-net. The LLF-w/o-nll has a slightly better UHD than LLF. We believe this is because, without using the likelihood, LLF-w/o-nll is trained directly by optimizing the Hausdorff distance. However, the sample diversity and quality, i.e., TMD and MMD, are worse than LLF. As argued by \citep{yang2019pointflow}, the current metrics for evaluating point cloud samples all have flaws, so these scores cannot be treated as hard metrics for evaluating model performance. We visualize some samples in Figure~\ref{fig:pc_sample1} and in appendix. These samples show that LLF can generate samples that comparable to other baselines.

% \begin{table*}[!htp]
% \vspace{-0.3cm}
% 	\begin{center}
% 		\caption{Evaluation results on three classes of PartNet. LLF performs comparable to baselines. }
% 		\label{tab:result_pc}
% 		\footnotesize
% 		\begin{tabular}{ l | l |  l  l  l l l}
% 		\toprule
% 		~ & ~ & LLF & LLF-w/o-nll & pcl2pcl & mm-pcl2pcl & mm-pcl2pcl-im \\
% 		\midrule
% 		\multirowcell{3}{MMD \\ (lower is better)} & Chair & $1.72$ & $1.79$ & $1.90$ & $\mathbf{1.52}$ & $1.90$ \\
% 		& Lamp & $2.11$ & $2.21$ & $2.50$ & $\mathbf{1.97}$ & $2.55$ \\
% 		& Table & $1.57$ & $1.57$ & $1.90$ & $\mathbf{1.46}$ & $1.54$ \\
% 		\midrule
% 		\multirowcell{3}{TMD \\ (higher is better)} & Chair & $0.63$ & $0.47$ & $0.00$ & $\mathbf{2.75}$ & $1.01$ \\
% 		& Lamp & $0.57$ & $0.41$ & $0.00$ & $\mathbf{3.31}$ & $0.56$ \\
% 		& Table & $0.55$ & $0.43$ & $0.00$ & $\mathbf{3.30}$ & $0.51$ \\
% 		\midrule
% 		\multirowcell{3}{UHD \\ (lower is better)} & Chair & $5.74$ & $5.49$ & $\mathbf{4.88}$ & $6.89$ & $6.65$ \\
% 		& Lamp & $4.71$ & $\mathbf{4.61}$ & $4.64$ & $5.72$ & $5.40$ \\
% 		& Table & $5.42$ & $5.13$ & $\mathbf{4.78}$ & $5.56$ & $5.38$ \\
% 		\bottomrule
% 		\end{tabular}
% 	\end{center}
% \end{table*}
% \vspace{-0.6cm}

\section{Conclusion}
\label{sec:conclusion}

In this paper, we propose label learning flows, which represent a general framework for weakly supervised learning. LLF uses a conditional flow to define the conditional distribution $p(\mathbf{y}|\mathbf{x})$, so that can model the uncertainty between input $\mathbf{x}$ and all possible $\mathbf{y}$. Learning LLF is a constrained optimization problem that optimizes the likelihood of possible $\mathbf{y}$ within the constrained space defined by weak signals. We develop a training method to train LLF inversely, avoiding the need of estimating $\mathbf{y}$. We apply LLF to three weakly supervised learning problems, and the results show that our method outperforms many state-of-the-art baselines on the weakly supervised classification and regression problems, and performs comparably to other new methods for unpaired point cloud completion. These results indicate that LLF is a powerful and effective tool for weakly supervised learning problems.

\section*{Acknowledgments}
We thank NVIDIA’s GPU Grant Program for its support. Wenzhuo Song is supported by the National Natural Science Foundation of China under Grant 62307006, and the Fundamental Research Funds for the Central Universities, NENU and JLU. The work was completed while You Lu and Chidubem Arachie were affiliated with the Virginia Tech Department of Computer Science, and Bert Huang was affiliated with the Tufts University Department of Computer Science. 

\appendix
\section{Analysis of LLF}
\label{sec:app_proof}

\renewcommand{\theequation}{\Alph{section}.\arabic{equation}}
\renewcommand{\thetable}{\Alph{section}.\arabic{table}}
\renewcommand{\thefigure}{\Alph{section}.\arabic{figure}}
\setcounter{equation}{0}
\setcounter{table}{0}
\setcounter{figure}{0}

Let $\hat{\mathbf{y}}$ be the true label. We assume that each data point $\mathbf{x}_i$ only has one unique label $\hat{\mathbf{y}}_i$, so that $p_{\text{data}}(\mathbf{x}, \hat{\mathbf{y}}) = p_{\text{data}}(\mathbf{x})$. Let $q(\hat{\mathbf{y}}|\mathbf{x})$ be a certain model of $\hat{\mathbf{y}}$. Traditional supervised learning learns a $q(\hat{\mathbf{y}}|\mathbf{x})$ that maximizes $\mathbb{E}_{p_{\text{data}}(\mathbf{x}, \hat{\mathbf{y}})}\left[\log q(\hat{\mathbf{y}}|\mathbf{x}, \phi)\right]$. 

Following theorem reveals the connection between LLF and dequantization~\citep{theis2015note,ho2019flow}, i.e., a commonly used technique for generative models that converts a discrete variable to continuous.

\begin{theorem}
	\label{theorem:lower_bound}
	Suppose that for any $i$, $\Omega^*_i$ satisfies that $\hat{\mathbf{y}}_i \in \Omega_i^*$, and for any two $\hat{\mathbf{y}}_i \not= \hat{\mathbf{y}}_j$, the $\Omega^*_i$ and $\Omega^*_j$ are disjoint. The volume of each $\Omega^*_i$ is bounded such that $\frac{1}{|\Omega^*_i|} \le M$, where $M$ is a constant. The relationship between $p(\mathbf{y}|\mathbf{x})$ and $q(\hat{\mathbf{y}}|\mathbf{x})$ can be defined as: $q(\hat{\mathbf{y}}|\mathbf{x}) = \int_{\mathbf{y} \in \Omega^*} p(\mathbf{y}|\mathbf{x}) d\mathbf{y}$. Then maximizing $\log p(\mathbf{y}|\mathbf{x})$ can be interpreted as maximizing the lower bound of $\log q(\hat{\mathbf{y}}|\mathbf{x})$. That is,
	\begin{align}
	&\mathbb{E}_{p_{\text{data}}(\mathbf{x})} \mathbb{E}_{\mathbf{y} \sim U(\Omega^*)}\left[\log p(\mathbf{y}|\mathbf{x}, \phi)\right] \nonumber\\
     &\le M \mathbb{E}_{p_{\text{data}}(\mathbf{x}, \hat{\mathbf{y}})}\left[\log q(\hat{\mathbf{y}}|\mathbf{x}, \phi)\right]
	\end{align}
\end{theorem}

The proof of Theorem~1 is similar to the proof of dequantization~\citep{theis2015note,ho2019flow}.

\noindent\textit{Proof.} 
\begin{align}
	\mathbb{E}_{p_{data}(\mathbf{x})}& \mathbb{E}_{\mathbf{y} \sim U(\Omega^*)}\left[\log p(\mathbf{y}|\mathbf{x}, \phi)\right]\\
        &= \sum_{\mathbf{x}} p_{data}(\mathbf{x}) \int_{\mathbf{y} \in \Omega^*} \frac{1}{|\Omega^*|}\log p(\mathbf{y}|\mathbf{x})d\mathbf{y} \nonumber\\
	&\le M \sum_{\mathbf{x}} p_{data}(\mathbf{x}) \log \int_{\mathbf{y} \in \Omega^*} p(\mathbf{y}|\mathbf{x})d\mathbf{y} \nonumber\\
	&= M \sum_{\mathbf{x}, \hat{\mathbf{y}}} p_{data}(\mathbf{x}, \hat{\mathbf{y}}) \log q(\hat{\mathbf{y}}|\mathbf{x}) \nonumber\\
	&= M \mathbb{E}_{p_{data}(\mathbf{x}, \hat{\mathbf{y}})}\left[\log q(\hat{\mathbf{y}}|\mathbf{x})\right]
\end{align}
In the first row, we expand the two expectations based on their definitions. In the second row, we use the property that $\frac{1}{|\Omega^*|} \le M$, and the Jensen's inequality, i.e., the integral of logarithm is less than or equal to the logarithm of integral. In the third row, we use the assumption that $p_{\text{data}}(\mathbf{x}) = p_{\text{data}}(\mathbf{x}, \mathbf{\hat{y}})$, and the relationship that $q(\hat{\mathbf{y}}|\mathbf{x}) = \int_{\mathbf{y} \in \Omega^*} p(\mathbf{y}|\mathbf{x}) d\mathbf{y}$.
\hfill $\square$

\vspace{0.2in}

Based on the theorem, when the constrained space is $\Omega^*$, learning with Eq. 3 in our paper is analogous to dequantization. That is, our method optimizes the likelihood of dequantized true labels. Maximizing $\log p(\mathbf{y}|\mathbf{x})$ can be interpreted as maximizing a lower bound of $\log q(\hat{\mathbf{y}}|\mathbf{x})$. Optimizing Eq. 3 in our paper will also optimize the certain model on true labels. In practice, the real constrained space, may not fulfill the assumptions for Theorem~\ref{theorem:lower_bound}, e.g., for some samples, the true label $\hat{\mathbf{y}}_i$ is not contained in the $\Omega_i$, or the constrains are too loose, so the $\Omega$s of different samples are overlapped. These will result in inevitable errors that come from the weakly supervised setting. Besides, for some regression problems, the ideal $\Omega^*$ only contains a single point: the ground truth label, i.e., $\forall i, \Omega^*_i = \{ \hat{\mathbf{y}}_i \}$.

% \section{LABEL LEARNING FLOW FOR UNPAIRED POINT CLOUD COMPLETION}
\section{Label Learning Flow for Unpaired Point Cloud Completion}

\renewcommand{\theequation}{\Alph{section}.\arabic{equation}}
\renewcommand{\thetable}{\Alph{section}.\arabic{table}}
\renewcommand{\thefigure}{\Alph{section}.\arabic{figure}}
\setcounter{equation}{0}
\setcounter{table}{0}
\setcounter{figure}{0}

In this section, we provide complete derivations of LLF for unpaired point cloud completion. The conditional likelihood $\log p(\mathbf{y}|\mathbf{x}_p)$ is an exchangeable distribution. We use De Finetti’s representation theorem and variational inference to derive a tractable lower bound for it.

\begin{align}
\log p(\mathbf{y} | \mathbf{x}_p) &= \int p(\mathbf{y}, \mathbf{u} | \mathbf{x}_p) d\mathbf{u} \nonumber\\
&= \int p(\mathbf{y} | \mathbf{u}, \mathbf{x}_p) p(\mathbf{u}) d\mathbf{u} \nonumber\\
&\ge \mathbb{E}_{q(\mathbf{u}|\mathbf{x}_p)} \left[ \log p(\mathbf{y} | \mathbf{u}, \mathbf{x}_p)\right] \nonumber\\
& \quad- \text{KL}(q(\mathbf{u}|\mathbf{x}_p) || p(\mathbf{u})) \nonumber\\
&\ge  \mathbb{E}_{q(\mathbf{u}|\mathbf{x}_p)} \left[ \sum_{i=1}^{T_c} \log p(\mathbf{y}_{i} | \mathbf{u}, \mathbf{x}_p)\right] \nonumber\\
&\quad - \text{KL}(q(\mathbf{u}|\mathbf{x}_p) || p(\mathbf{u})),
\end{align}
where in the third row, we use Jensen's inequality to compute the lower bound, and in the last row, we use De Finetti's theorem to factorize $p(\mathbf{y}|\mathbf{u},\mathbf{x}_p)$ to the distributions of points.

The least square GAN discriminator and the Hausdorff distance for generated complete point clouds can be treated as two equality constraints
\begin{align*}
    & D(\mathbf{y}) = 1 \\
    & d_H(\mathbf{y}, \mathbf{x}_p) = 0.
\end{align*}

Note that the $d_H()$ is non-negative. Convert these two constraints to penalty functions, we have
\begin{align}
\label{eq:point_cloud_obj_deri}
&\max_{\phi} \mathbb{E}_{q(\mathbf{u}|\mathbf{x}_p)} \left[ \sum_{t=1}^{T_c} \log p_Z(\mathbf{z_t}) \right. \nonumber\\
& \left. - \sum_{i=1}^{K} \log \left|\det \left(\frac{\partial \mathbf{g}_{\mathbf{u}, \mathbf{x}_p, \phi_i}}{\partial \mathbf{r}_{t,i}}\right)\right|\right]\nonumber\\
& - \mathbb{E}_{q(\mathbf{u}|\mathbf{x}_p)} \left[ \lambda_1 (D(\mathbf{g}_{\mathbf{u}, \mathbf{x}_p,\phi}(\mathbf{z})) -1)^2 \right. \nonumber\\
& \left. + \lambda_2 d^{HL}(\mathbf{g}_{\mathbf{u}, \mathbf{x}_p, \phi}(z), \mathbf{x}_p) \right] \nonumber\\ 
& - \text{KL}(q(\mathbf{u}|\mathbf{x}_p) || p(\mathbf{u})).
\end{align}

% \section{EXPERIMENT DETAILS}
\section{Experiment Details}
\label{app_sec:experiment_details}

\renewcommand{\theequation}{\Alph{section}.\arabic{equation}}
\renewcommand{\thetable}{\Alph{section}.\arabic{table}}
\renewcommand{\thefigure}{\Alph{section}.\arabic{figure}}
\setcounter{equation}{0}
\setcounter{table}{0}
\setcounter{figure}{0}

In this section, we provide more details on our experiments to help readers reproduce our results.

\subsection{Model Architectures}

For experiments of weakly supervised classification, and unpaired point cloud completion, we use normalizing flows with only conditional affine coupling layers~\citep{klokov2020discrete}. Each layer is defined as 
\begin{align}
\label{eq:cond_affine_dpf}
& \mathbf{y}_a, \mathbf{y}_b = \text{split}(\mathbf{y}) \nonumber\\
& \mathbf{s} = \mathbf{m}_s(\mathbf{w}_y(\mathbf{y}_a) \odot \mathbf{w}_x(\mathbf{x}) + \mathbf{w}_b(\mathbf{x})) \nonumber\\
& \mathbf{b} = \mathbf{m}_b(\mathbf{c}_y(\mathbf{y}_a) \odot \mathbf{c}_x(\mathbf{x}) + \mathbf{c}_b(\mathbf{x})) \nonumber\\
& \mathbf{z}_b = \mathbf{s}\odot\mathbf{y}_b + \mathbf{b} \nonumber\\
& \mathbf{z} = \text{concat}(\mathbf{y}_a, \mathbf{z}_b),
\end{align}
where $\mathbf{m}, \mathbf{w}, \mathbf{c}$ are all small neural networks.

For weakly supervised regression, since the label $y$ is a scalar, we use conditional affine transformation as a flow layer, which is defined as
\begin{equation}
\label{eq:cond_affine_trans}
    y = \mathbf{s}(\mathbf{x})*z + \mathbf{b}(\mathbf{x})
\end{equation}
where $\mathbf{s}$ and $\mathbf{b}$ are two neural networks that take $\mathbf{x}$ as input and output parameters for $y$.

For LLF, we only need the inverse flow, i.e., $\mathbf{g}_{\mathbf{x},\phi}$, for training and prediction, so in our experiments, we actually define $\mathbf{g}_{\mathbf{x},\phi}$ as the forward transformation, and let $\mathbf{y} = \mathbf{s}\odot \mathbf{z} + \mathbf{b}$. We do this because multiplication and addition are more stable than division and subtraction. 

\subsubsection{Weakly Supervised Classification} 

In this problem, we use a flow with $8$ flow steps, and each step has $2$ conditional affine coupling layers. These two layers will transform different dimensions. Each $\mathbf{w}$ and $\mathbf{c}$ are small MLPs with two linear layers. Each $\mathbf{m}$ has one linear layer. The hidden dimension of linear layers is fixed to $64$.

\subsubsection{Weakly Supervised Regression}

In this problem, we use conditional affine transformation introduced in Eq.~\ref{eq:cond_affine_trans}, as a flow layer. A flow has $8$ flow layers. The $\mathbf{s}$ and $\mathbf{b}$ in a flow layer are three layer MLPs. The hidden dimension of linear layers is $64$.

\subsubsection{Unpaired Point Cloud Completion}

\begin{figure}[!htp]
\vspace{-0.5cm}
	\centering
   \includegraphics[width=0.8\linewidth]{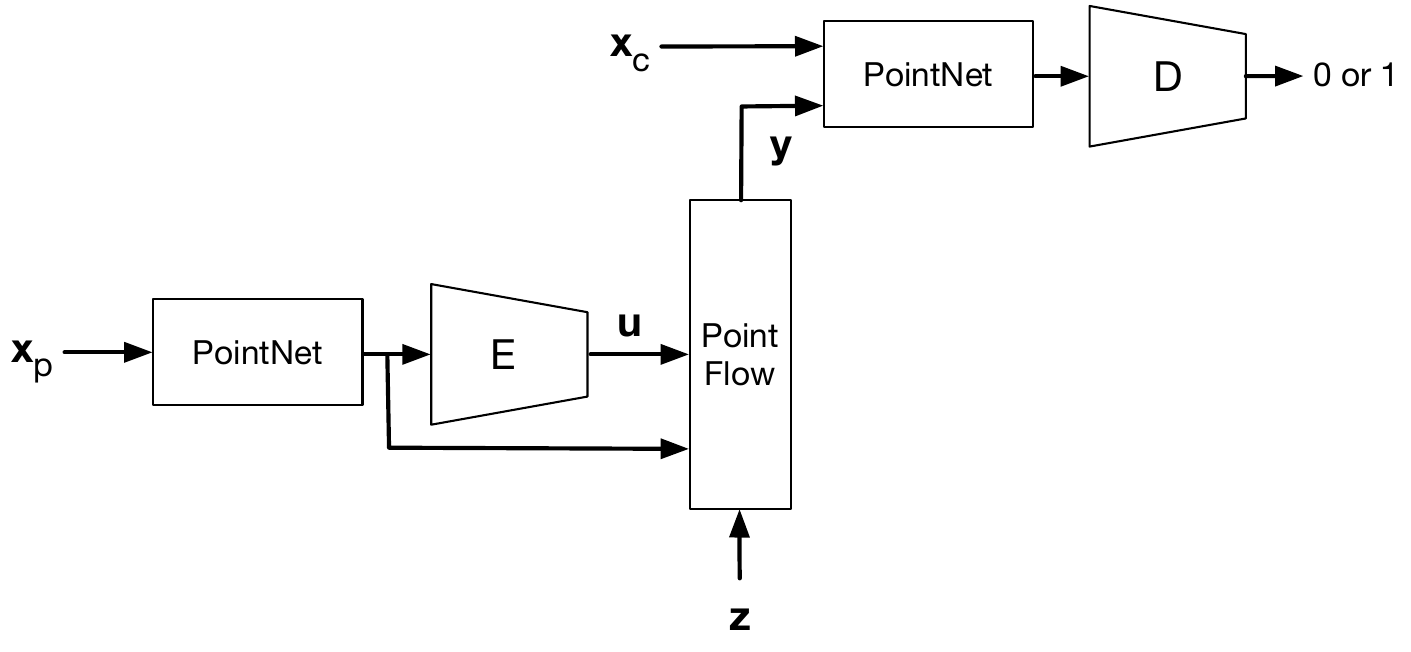}
\caption{Model architecture of LLF for unpaired point cloud completion. The $E$ represents the encoder, and the $D$ represents the GAN discriminator.}
\label{fig:llf-architecture}
	%\vspace{-0.1in}
\end{figure}

The model architecture used LLF used for this problem is illustrated in Figure~\ref{fig:llf-architecture}. We use the same architecture as DPF~\citep{klokov2020discrete} for point flow. Specifically, the flow has $8$ flow steps, and each step has 3 conditional affine coupling layers, i.e., Eq.~\ref{eq:cond_affine_dpf}. Slightly different from the original DPF, the conditioning networks $\mathbf{w}_x$, $\mathbf{c}_x$, $\mathbf{w}_b$, and $\mathbf{c}_b$ will take the latent variable $\mathbf{u}$ and the features of partial point cloud $\mathbf{x}_p$ as input. The $\mathbf{w}$s and $\mathbf{c}$s are MLPs with two linear layers, whose hidden dimension is $64$. The $\mathbf{m}$s are one layer MLPs.

We use a PointNet~\citep{qi2017pointnet} to extract features from partial point cloud $\mathbf{x}_p$. Following \citep{klokov2020discrete}, the hidden dimensions of this PointNet is set as $64-128-256-512$. Given the features of $\mathbf{x}_p$, the encoder $E$ then uses the reparameterization trick~\citep{kingma2013auto} to generate latent variable $\mathbf{u}$, which is a $128$-dimensional vector. The encoder has two linear layers, whose hidden dimension is $512$.

The GAN discriminator uses another PointNet to extract features from (generated) complete point clouds. We follow \citep{wu2020multimodal} and set the hidden dimensions of this PointNet as $64-128-128-256-128$. The discriminator $D$ is a three layer MLP, whose hidden dimensions are $128-256-512$. 

\subsection{Experiment setup}
In weakly supervised classification and regression experiments, we fix $\lambda = 10$. and use default settings, i.e., $\eta = 0.001$, $\beta_1=0.9$ and $\beta_2=0.999$ for Adam~\citep{kingma2014adam}. We use an exponential learning rate scheduler with a decreasing rate of $0.996$ to guarantee convergence. We track the decrease of loss and when the decrease is small enough, the training stops. The random seeds we use in our experiments are $\{0,10,100,123,1234\}$. For experiments on tabular datasets and image sets, we set the maximum epochs to $2000$. For experiments on real text datasets, we set the maximum epochs to $500$.

For experiments with unpaired point cloud completion, we use Adam with an initial learning rate $\eta = 0.0001$ and default $\beta$s. The best coefficients for the constraints in Eq.~12 of main paper are $\lambda_1=10, \lambda_2=100$. We use stochastic optimization to train the models, and the batch size is $32$. Each model is trained for at most $2000$ epochs. 

All experiments are conducted with $1$ GPU.

\subsection{Data, Training, and Evaluation Details}

\subsubsection{Weakly Supervised Classification}

\begin{table*}[htp]
\vspace{-0.5cm}
	\begin{center}
		\caption{Summary of datasets used in weakly supervised classification experiments. The ``---" indicates this dataset does not have a official split}
		\label{tab:dataset_classification}
		\begin{tabular}{l l l l l l }
			\toprule
			Dataset & Size & Train Size & Test Size & No. features & No. weak signals  \\
			\midrule
			Fashion MNIST (DvK) & $14,000$ & $12,000$ & $2,000$ & $784$ & $3$ \\
			Fashion MNIST (SvA) & $14,000$ & $12,000$ & $2,000$ & $784$ & $3$ \\
			Fashion MNIST (CvB) & $14,000$ & $12,000$ & $2,000$ & $784$ & $3$ \\
			Breast Cancer & $569$ & --- & --- & $30$ & $3$ \\
			OBS Network & $795$ & --- & --- & $21$ & $3$ \\
			Cardiotocography & $963$ & --- & --- & $21$ & $3$ \\
			Clave Direction & $8,606$ & --- & --- & $16$ & $3$ \\
			Credit Card & $1,000$ & --- & --- & $24$ & $3$ \\
			Statlog Satellite & $3,041$ & --- & --- & $36$ & $3$ \\
			Phishing Websites & $11,055$ & --- & --- & $30$ & $3$ \\
			Wine Quality & $4,974$ & --- & --- & $11$ & $3$ \\
			IMDB & $49,574$ & $29,182$ & $20,392$ & $300$ & $10$ \\
			SST & $5,819$ & $3,998$ & $1,821$ & $300$ & $14$ \\
			YELP & $55,370$ & $45,370$ & $10,000$ & $300$ & $14$ \\
			\bottomrule
		\end{tabular}
	\end{center}
\end{table*}
% \begin{table*}[htp]
% \vspace{-0.5cm}
% 	\begin{center}
% 		\caption{Summary of datasets used in weakly supervised regression experiments}
% 		\label{tab:dataset_regression}
% 		\begin{tabular}{l l l l }
% 			\toprule
% 			Dataset & Size & No. features & No. weak signals  \\
% 			\midrule
% 			Air Quality & $8,991$ & $12$ & $5$ \\
% 			Temperature Forecast & $7,588$ & $24$ & $5$ \\
% 			Bike Sharing & $17,379$ & $12$ & $5$\\
% 			\bottomrule
% 		\end{tabular}
% 	\end{center}
% \end{table*}
For experiments on tabular and image datasets, we use the same approach as \citep{arachie2021general,arachie2020constrained} to split each dataset to training, simulation, and test sets. We use the data and labels in simulation sets to create weak signals, and estimated bounds. We train models on training sets and test model on test sets. We assume that the models do not have access to any labels. The labels in simulation sets are only used to generated weak signals and estimate bounds. We follow \citep{arachie2021general} and choose $3$ features to create weak signals. We train a logistic regression with each feature on the simulation set, and use the label probabilities predicted by this logistic regression as weak signals. We compute the error of this trained logistic regression on simulation set as estimated error bound. Note that the weak signals of each data point are probabilities that this sample belongs to the positive class. For PGMV~\citep{mazzetto2021semi} and ACML~\citep{mazzetto:icml21}, we round the probabilities to form one-hot vectors.

For experiments on real text datasets, we use the same keyword-based method as \citep{arachie2020constrained} to create  weak supervision. Specifically, we choose key words that can weakly indicate positive and negative sentiments. Documents containing positive words will be labeled as positive, and vice versa. The weak signals in this task are one-hot vectors, indicating which class a data point belongs to. If a key word is missing in a document, the corresponding weak label will be null. For two-stage methods, we follow \citep{arachie2020constrained} and use a two layer MLP as the classifier, whose latent dimension is $512$.

 We list some main features of these datasets in Table~\ref{tab:dataset_classification}. We refer readers to their original papers for more details. For those datasets without official splits, we randomly split them with a ratio of $4:3:3$. 

% \begin{table*}[htp]
% 	\begin{center}
% 		\caption{Summary of datasets used in weakly supervised classification experiments. The ``---" indicates this dataset does not have a official split}
% 		\label{tab:dataset_classification}
% 		\begin{tabular}{l l l l l l }
% 			\toprule
% 			Dataset & Size & Train Size & Test Size & No. features & No. weak signals  \\
% 			\midrule
% 			Fashion MNIST (DvK) & $14,000$ & $12,000$ & $2,000$ & $784$ & $3$ \\
% 			Fashion MNIST (SvA) & $14,000$ & $12,000$ & $2,000$ & $784$ & $3$ \\
% 			Fashion MNIST (CvB) & $14,000$ & $12,000$ & $2,000$ & $784$ & $3$ \\
% 			Breast Cancer & $569$ & --- & --- & $30$ & $3$ \\
% 			OBS Network & $795$ & --- & --- & $21$ & $3$ \\
% 			Cardiotocography & $963$ & --- & --- & $21$ & $3$ \\
% 			Clave Direction & $8,606$ & --- & --- & $16$ & $3$ \\
% 			Credit Card & $1,000$ & --- & --- & $24$ & $3$ \\
% 			Statlog Satellite & $3,041$ & --- & --- & $36$ & $3$ \\
% 			Phishing Websites & $11,055$ & --- & --- & $30$ & $3$ \\
% 			Wine Quality & $4,974$ & --- & --- & $11$ & $3$ \\
% 			IMDB & $49,574$ & $29,182$ & $20,392$ & $300$ & $10$ \\
% 			SST & $5,819$ & $3,998$ & $1,821$ & $300$ & $14$ \\
% 			YELP & $55,370$ & $45,370$ & $10,000$ & $300$ & $14$ \\
% 			\bottomrule
% 		\end{tabular}
% 	\end{center}
% \end{table*}

\subsubsection{Weakly Supervised Regression}

We use three datasets from the UCI repository. For each dataset, we randomly split it to training, simulation, and test sets with a ratio of $4:3:3$. We use the simulation set to create weak signals and estimated label values. We choose $5$ features to create weak signals for each dataset. The detailed introduction of these datasets are as follows. Table~\ref{tab:dataset_regression} summarize the statistical results of them. 

\textbf{Air Quality.} In this task, we predict the absolute humidity in air, based on other air quality features such as hourly averaged temperature, hourly averaged $\text{NO}_2$ concentration etc. The raw dataset has $9,358$ instances. We remove those instances with Nan values, resulting in a dataset with $8,991$ instances. We use hourly averaged concentration CO, hourly averaged Benzene concentration, hourly averaged $\text{NO}_{\text{x}}$ concentration, tungsten oxide hourly averaged sensor response, and relative humidity as features for creating weak signals.

\textbf{Temperature Forecast.} In this task, we predict the next day maximum air temperature based on current day information. The raw dataset has $7,750$ instances, and we remove those instances with Nan values, resulting in $7,588$ instances. We use present max temperature, forecasting next day wind speed, forecasting next day cloud cover, forecasting next day precipitation, solar radiation as features for creating weak signals.

\textbf{Bike Sharing.} In this task, we predict the count of total rental bikes given weather and date information. The raw dataset has $17,389$ instances, and we remove those instances with Nan values, resulting in $17,379$ instances. We use season, hour, if is working day, normalized feeling temperature, and wind speed as features for creating weak signals.
%%%%%%%%%%%%%%%%%%%%%%%%%%%%
\begin{table*}[htp]
\vspace{-0.5cm}
	\begin{center}
		\caption{Summary of datasets used in weakly supervised regression experiments}
		\label{tab:dataset_regression}
		\begin{tabular}{l l l l }
			\toprule
			Dataset & Size & No. features & No. weak signals  \\
			\midrule
			Air Quality & $8,991$ & $12$ & $5$ \\
			Temperature Forecast & $7,588$ & $24$ & $5$ \\
			Bike Sharing & $17,379$ & $12$ & $5$\\
			\bottomrule
		\end{tabular}
	\end{center}
\end{table*}

In these datasets, the original label is within an interval $[l_y, u_y]$. In training, we normalize the original label to within $[0,1]$ by computing $y = (y - l_y)/ (u_y-l_y)$. In prediction, we recover the predicted label to original value by computing $y = y(u_y-l_y)+l_y$.
\begin{table}[htp]

	\begin{center}
		\caption{Summary of datasets used unpaird point cloud completion}
		\label{tab:dataset_pc}
		\begin{tabular}{l l l l }
			\toprule
			Dataset & Train Size & Valid Size & Test Size  \\
			\midrule
			Chair & $4,489$ & $617$ & $1,217$ \\
			Table & $5,707$ & $843$ & $1,668$ \\
			Lamp & $1,545$ & $234$ & $416$\\
			\bottomrule
		\end{tabular}
	\end{center}
\end{table}
\subsubsection{Unpaired Point Cloud Completion}

\textbf{Datasets.} We use the same way as \citep{wu2020multimodal} to process PartNet~\citep{mo2019partnet}. PartNet provides point-wise semantic labels for point clouds. The original point clouds are used as complete point clouds. To generate partial point clouds, we randomly removed parts from complete point clouds, based on the semantic labels. We use Chair, Table, and Lamp categories. The summary of these three subsets are in Table~\ref{tab:dataset_pc}.
%%%%%%%%%%%%%%%%
% \begin{table}[hp]
% \vspace{-0.5cm}
% 	\begin{center}
% 		\caption{Summary of datasets used unpaird point cloud completion experiments}
% 		\label{tab:dataset_pc}
% 		\begin{tabular}{l l l l }
% 			\toprule
% 			Dataset & Train Size & Valid Size & Test Size  \\
% 			\midrule
% 			Chair & $4,489$ & $617$ & $1,217$ \\
% 			Table & $5,707$ & $843$ & $1,668$ \\
% 			Lamp & $1,545$ & $234$ & $416$\\
% 			\bottomrule
% 		\end{tabular}
% 	\end{center}
% \end{table}

\textbf{Metrics.} Let $\mathcal{X}_c$ be the set of referred complete point clouds, and $\mathcal{X}_p$ be the set of input partial point clouds. For each partial point cloud $\mathbf{x}^{(p)}_i$, we generate $M$ complete point cloud samples $\mathbf{y}_{i}^{(1)},...,\mathbf{y}_{i}^{(M)}$. All these samples form a new set of complete point clouds $\mathcal{Y}$. In our experiments, we follow \citep{wu2020multimodal} and set $M=10$.

The MMD~\citep{achlioptas2018learning} is defined as
\begin{equation}
    \text{MMD} = \frac{1}{|\mathcal{X}_c|} \sum_{\mathbf{x}_i \in \mathcal{X}_c}d_{C}(\mathbf{x}_i, \text{NN}(\mathbf{x}_i)),
\end{equation}
where $\text{NN}(\mathbf{x})$ is the nearest neighbor of $\mathbf{x}$ in $\mathcal{Y}$. The $d_C$ represents Chamfer distance. MMD computes the distance between the set of generated samples and the set of target complete shapes, so it measures the quality of generated.

The TMD is defined as
\begin{equation}
    \text{TMD} = \frac{1}{|\mathcal{X}_p|} \sum_{i=1}^{|\mathcal{X}_p|} \left( \frac{2}{M-1} \sum_{j=1}^{M}\sum_{k=j+1}^M d_{C}(\mathbf{y}_i^{(j)}, \mathbf{y}_i^{(k)}) \right).
\end{equation}
TMD measures the difference of generated samples given an input partial point cloud, so it measures the diversity of samples.

The UHD is defined as
\begin{equation}
    \text{UHD} = \frac{1}{|\mathcal{X}_p|} \sum_{i=1}^{|\mathcal{X}_p|} \left(\frac{1}{M} \sum_{j=1}^M d_H(\mathbf{x}_i, \mathbf{y}_i^{(j)})\right),
\end{equation}
where $d_H$ represents the unidirectional Hausdorff distance. UHD measures the similarity between samples and input partial point clouds, so it measures the fidelity of samples.

\subsubsection{Running Time Comparison}
For weakly classification and regression, the training of LLF is slower than other baselines. This is mainly because LLF uses a deep learning model. Other baselines use non-deep models or small neural networks, e.g., ALL uses a 5 layer MLP. However, the training of LLF is still fast. We train LLF on 1 GPU, and the average training time is around $1,000$ seconds per task. For inference, LLF is as fast as other methods. For example, LLF can make predictions for 2000 samples in the Fashion MNIST test set in 0.3s, which is comparable to other methods.

For point cloud completion, LLF is as fast as mm-pcl2pcl, since they are all deep learning models, and their model sizes are close. They require around 60s to complete 1 epoch of training. In terms of inference, generating 1 sample takes around 0.05s. The experiments are run on 1 GPU.
We did not experience gradient explosion problems during training of LLF. We feel that this is mainly because we don’t need huge flow models, and complicated flow layers.

\subsubsection{Point Cloud Samples}
\begin{figure}[!htp]
\vspace{0.5cm}
	\centering
   \includegraphics[width=1.0\linewidth]{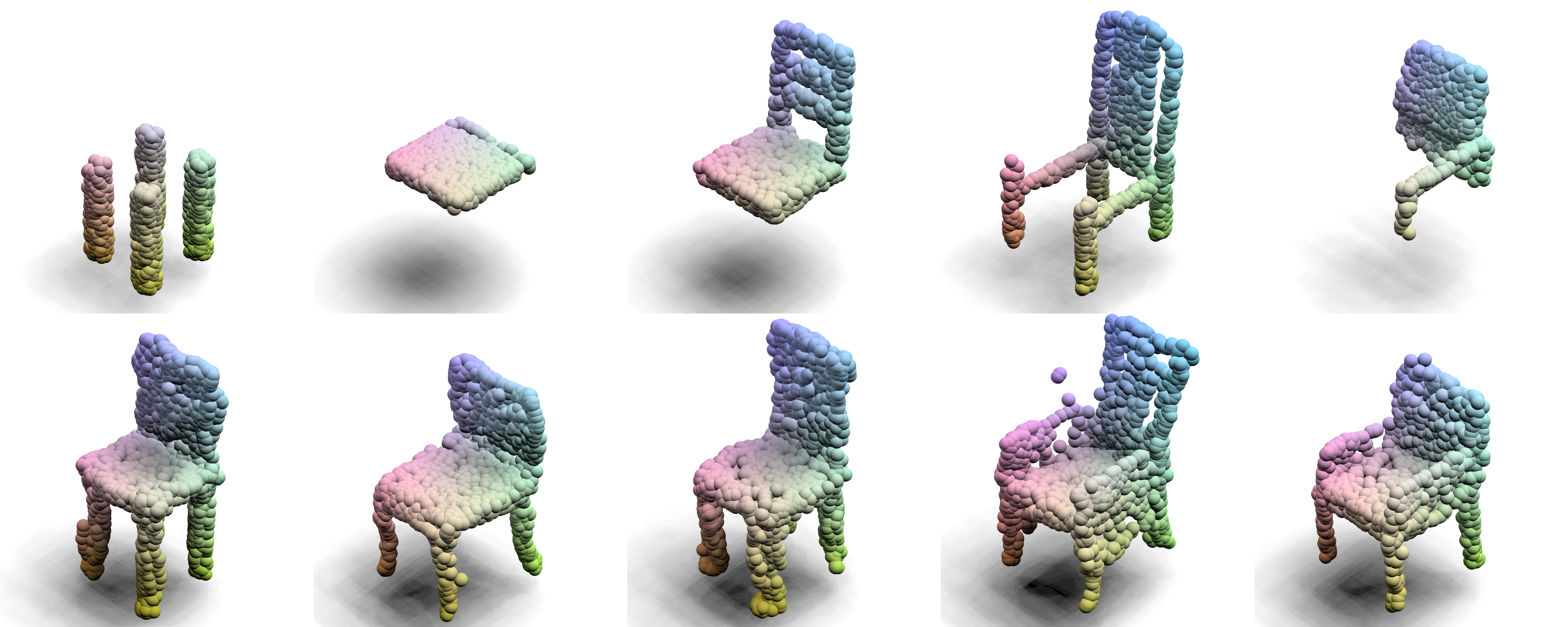}
\caption{Random chair samples generated by LLF. The first row is partial point clouds, and the second row is generated complete point clouds.}
\label{fig:pc_sample_chair}
	%\vspace{-0.1in}
\end{figure}

\begin{figure}[!htp]
\vspace{0.1cm}
	\centering
   \includegraphics[width=1.0\linewidth]{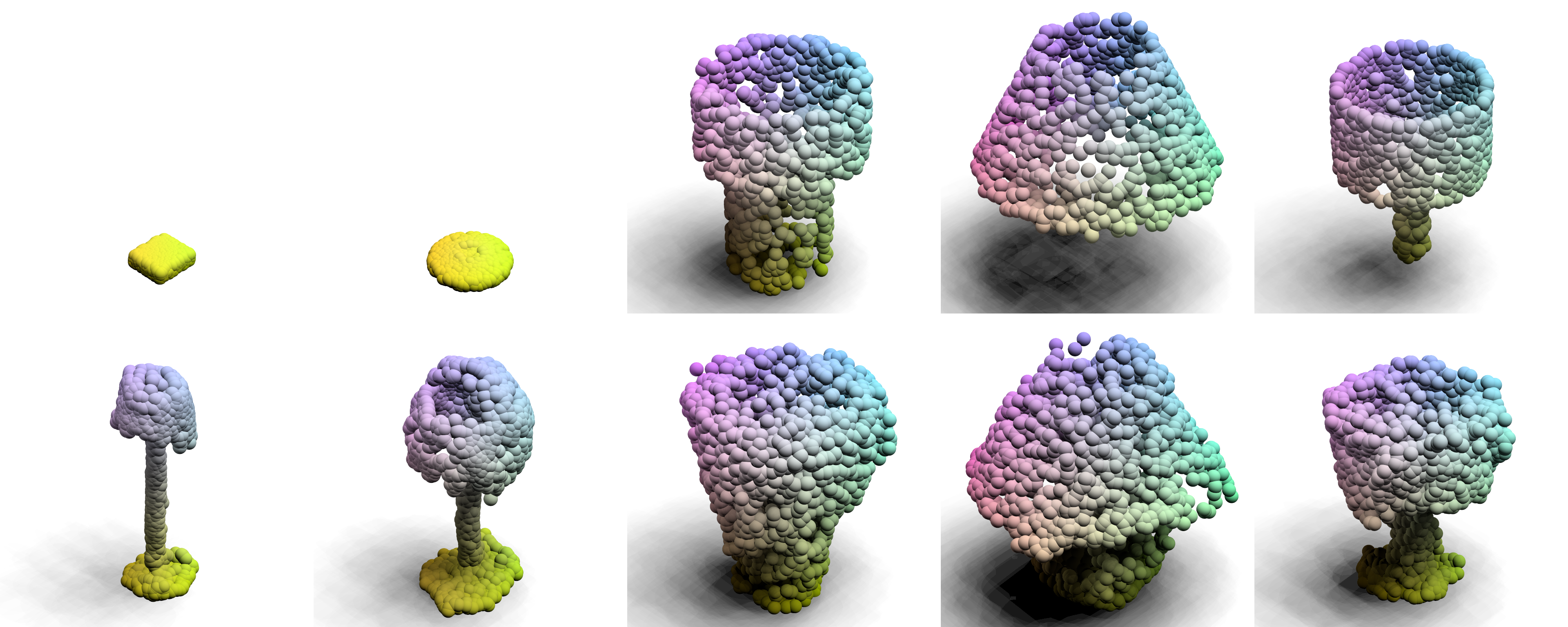}
\caption{Random lamp samples generated by LLF. }
\label{fig:pc_sample_lamp}
	%\vspace{-0.1in}
\end{figure}

\begin{figure}[!htp]
\vspace{0.1cm}
	\centering
   \includegraphics[width=1.0\linewidth]{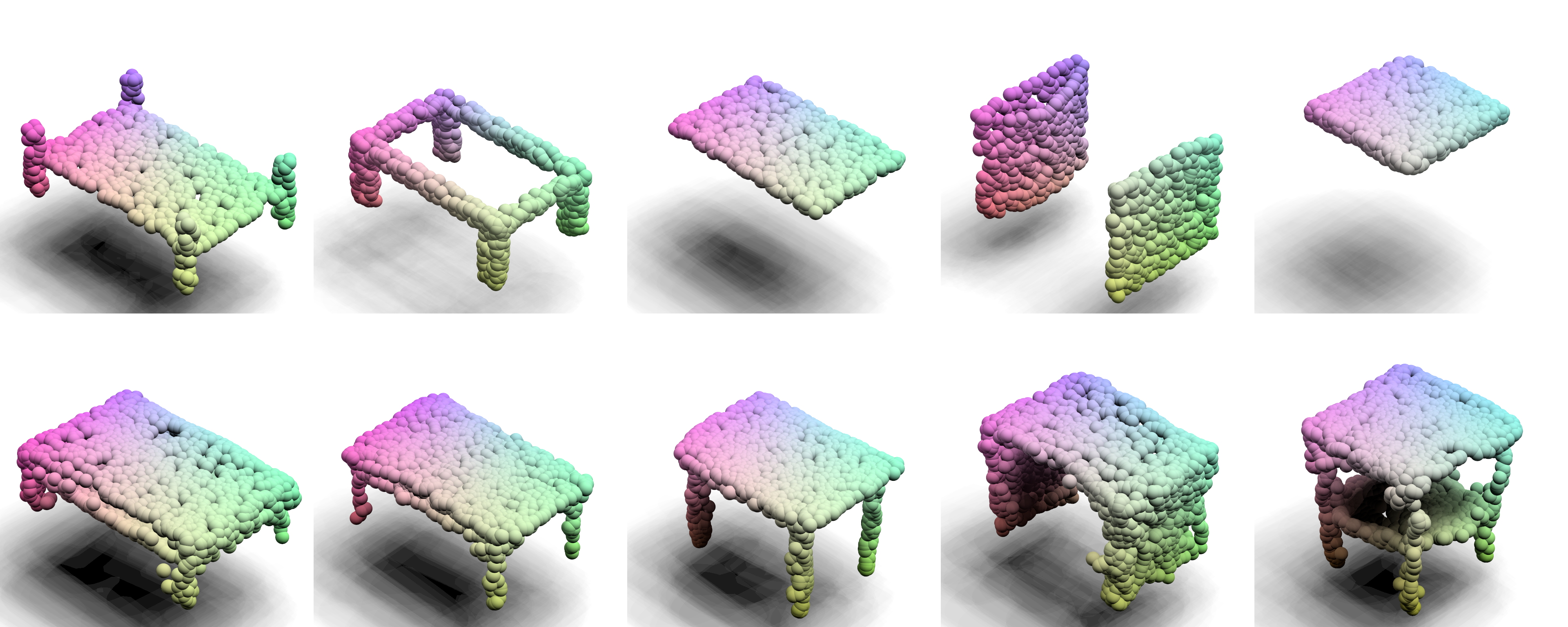}
\caption{Random table samples generated by LLF. }
\label{fig:pc_sample_table}
	%\vspace{-0.1in}
\end{figure}

We show more samples of LLF in Figure~\ref{fig:pc_sample_chair}, ~\ref{fig:pc_sample_lamp}, and ~\ref{fig:pc_sample_table}.

% \section{Societal Impact}
% One potential danger of this research is that weakly supervised learning may be applied to data with private information, e.g., clinical data. Due to privacy protections, labeling these data points are impossible. Weakly supervised learning is a possible way to circumvent these restrictions.

%\newpage

\printcredits


% \bibliographystyle{cas-model2-names}


% \bibliography{youlu}

\end{document}